\pdfoutput=1

\documentclass{article} %
\usepackage{iclr2027_conference,times}

\usepackage{amsmath,amsfonts,bm}

\def\eqref#1{equation~\ref{#1}}

\def\1{\bm{1}}

\DeclareMathAlphabet{\mathsfit}{\encodingdefault}{\sfdefault}{m}{sl}
\SetMathAlphabet{\mathsfit}{bold}{\encodingdefault}{\sfdefault}{bx}{n}

\usepackage{hyperref}
\usepackage{url}
\usepackage{graphicx}
\usepackage{float}
\usepackage{xcolor}
\definecolor{ForestGreen}{RGB}{34,139,34}
\usepackage{colortbl}
\usepackage{xspace}
\usepackage{booktabs}
\usepackage{tabularx}
\usepackage{amssymb}
\usepackage{amsmath}
\usepackage{multirow}
\usepackage{enumitem}
\usepackage{listings}
\usepackage[most]{tcolorbox}
\lstdefinestyle{prompt}{basicstyle=\ttfamily\scriptsize, breaklines=true, breakatwhitespace=false, columns=fullflexible, keepspaces=true, showstringspaces=false, breakindent=1em, postbreak=\mbox{\textcolor{gray}{$\hookrightarrow$}\space}}
\definecolor{vidd}{RGB}{40,120,60}\definecolor{ehrd}{RGB}{180,90,20}\definecolor{crossd}{RGB}{50,70,160}
\newcommand{\correct}{\textcolor{vidd}{\checkmark}}
\newtcolorbox{mcqcard}[2]{enhanced, breakable, colback=white, colframe=#1, boxrule=0.5pt, arc=2pt, left=6pt, right=6pt, top=4pt, bottom=4pt, before skip=6pt, after skip=6pt, title={#2}, fonttitle=\bfseries\footnotesize, colbacktitle=#1, coltitle=white, fontupper=\footnotesize}
\newtcolorbox{promptbox}[2][]{enhanced, breakable, colback=gray!4, colframe=black!55, boxrule=0.4pt, arc=1.5pt, left=4pt, right=4pt, top=2pt, bottom=2pt, before skip=6pt, after skip=8pt, title={#2}, fonttitle=\bfseries\footnotesize, colbacktitle=black!70, coltitle=white, #1}

\newcommand{\method}{TRACER\xspace}
\newcommand{\mcqbenchmark}{TRAJECTORY-144\xspace}

\newif\ifpreprint
\preprinttrue
\ifdefined\buildpreprint\preprinttrue\fi
\ifpreprint
  \iclrfinalcopy
  \newcommand{\site}{Stanford Hospital\xspace}
  \newcommand{\siteirb}{the Stanford Institutional Review Board\xspace}
\else
  \newcommand{\site}{a large academic medical center\xspace}
  \newcommand{\siteirb}{the Institutional Review Board of the study institution\xspace}
\fi

\title{\method: A Multimodal Harness for \\Patient Care in Hospitals}

\ifpreprint
\author{%
    Zane~Durante$^{\dagger}$,
    Paul~Tang$^{\dagger}$,
    Keshigeyan~Chandrasegaran$^{\dagger}$,\\
    \bf Amit~Kaushal,
    \bf Shrinidhi~Kowshika~Lakshmikanth,
    Feolu~Kolawole,\\
    \bf June~(Yu~Jun)~Zheng,
    \bf Yifan~Geng,
    Shobhit~Agarwal,
    Kevin~Schulman,\\
    \bf Juan~Carlos~Niebles,
    \bf Ehsan~Adeli,
    Arnold~Milstein,
    Li~Fei-Fei\\[0.4em]
    {\normalfont Stanford University}\\[0.2em]
    {\normalfont\small $^{\dagger}$Equal contribution}
}
\else
\author{Anonymous authors\\Paper under double-blind review}
\fi

\begin{document}

\maketitle
\ifpreprint\lhead{Preprint. Under review.}\fi

\begin{abstract}

Continuous video and clinical records provide complementary information for understanding how patient behavior evolves over a hospital stay. A central challenge is how to represent behavior across the entire stay while preserving access to fine-grained visual and clinical evidence. We introduce \method{}, a training-free multimodal harness that constructs minute-level action records from video and combines them with clinical records into hourly reports. The harness equips an LLM agent with tools to query this hierarchy, inspect relevant video, and retrieve clinical records. To evaluate \method{}, we construct two benchmarks. The first contains 397 annotated one-minute clips spanning 15 atomic actions; the second poses 144 clinician-informed multiple-choice questions, grounded in 264 hours of video and clinical records from two ICU patients. \method{} achieves up to 84.8\% accuracy for action detection and 61.0\% QA accuracy versus 49.3\% for the strongest baseline using the same backbone. Across 280 patients spanning approximately 60,000 camera-hours, \method{} identifies significantly less purposeful activity in patients with a positive delirium assessment during their stay than in those without ($p < 0.0001$). Our results show that combining continuous video with clinical records can support understanding of patient behavior over time and across patient cohorts without training on protected health data.

\end{abstract}

\section{Introduction}

Reasoning over long, continuous streams of visual observations is important for understanding human behavior. Recent advances in long-form video understanding have improved vision-language models' (VLMs) ability to recognize, summarize, and reason about events and activities over increasingly long visual contexts~\citep{chandrasegaran2024hourvideo,fu2024videomme,wu2024longvideobench,zhang2025deepvideodiscovery,yang2025egolife}. However, understanding real-world human behavior often requires integrating visual observations with external information. This need is particularly important in healthcare, where behavioral changes provide evidence about a patient's condition but must be interpreted within the context of their medical history and ongoing treatment. In this work, we study this problem in the intensive care unit (ICU), a high-stakes environment where continuous video and clinical records together capture changes in patient movement, responsiveness, interaction, and care across days~\citep{devlin2018clinical}.

Developing this capability in the ICU presents several technical challenges. Long-context vision--language models and caption-based pipelines offer approaches for long-form video understanding~\citep{chandrasegaran2024hourvideo,zeng2022socratic}, but applying them across entire ICU stays requires balancing temporal coverage with access to detail. Specifically, a system must be able to connect brief movements and interactions to behavioral patterns extending across days. And while bedside video captures observable activity and interactions, electronic health records (EHRs) document clinical assessments and interventions that provide complementary behavioral context. Finally, acquiring continuous bedside recordings linked to EHRs requires dedicated sensing infrastructure and secure handling of sensitive patient information; the recordings and clinical records used in our study cannot be released publicly. We therefore seek a method for longitudinal understanding that can be shared without releasing protected patient records or model parameters fitted
to them.

\begin{figure}[t]
    \centering
    \includegraphics[width=0.98\linewidth]{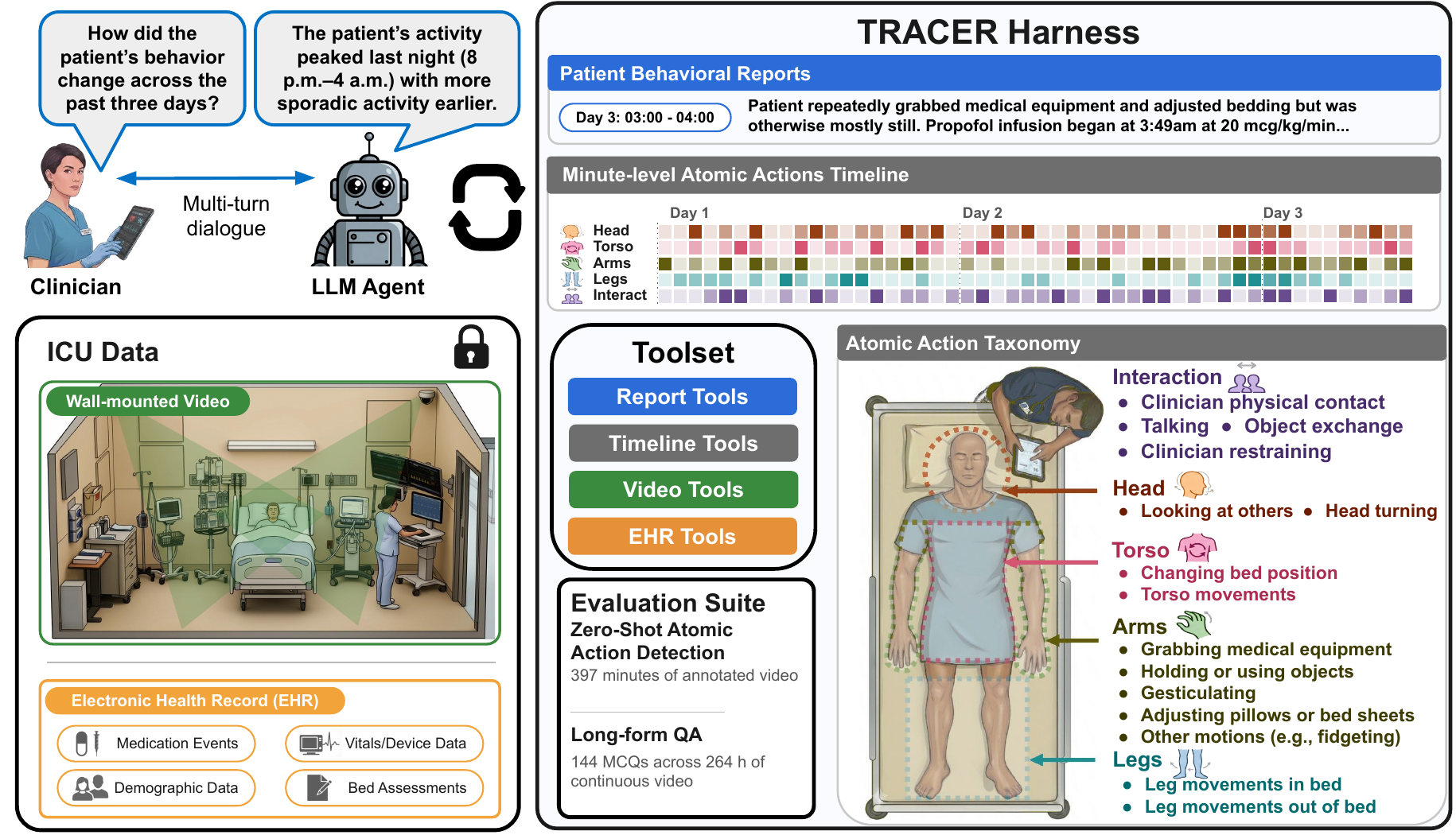}
    \caption{\textbf{Overview of \method.}
    TRACER equips an agent model with hierarchical behavioral representations and tools for understanding multi-day ICU patient trajectories. Vision–language models analyze one-minute bedside video segments to detect atomic actions defined by a clinically informed taxonomy and generate descriptive captions. These outputs are combined with concurrent medication records to generate structured hourly behavioral reports. Through report, timeline, video, and EHR tools, the agent iteratively retrieves evidence and examines relevant video segments to answer retrospective clinical questions. Our evaluation suite assesses atomic action detection on 397 annotated one-minute clips and longitudinal clinical reasoning on 144 multiple-choice questions spanning 264 hours of continuous video.}
    \label{fig:teaser}
\end{figure}

To address these challenges, we introduce \method{} (Trajectory Reasoning Across Clinical Event Records), a training-free multimodal harness for understanding multi-day ICU patient trajectories. \method{} separates temporal abstraction, which builds a hierarchy of minute- and hour-level records continuously as the stay is recorded, from evidence acquisition, in which an agent reasons over these compact records and accesses the raw video and EHR as needed. To construct this hierarchy, vision--language models (VLMs) analyze one-minute video segments zero-shot, producing atomic action detections under a clinically informed taxonomy together with descriptive captions. A language model combines these outputs with concurrent medication records to generate structured hourly behavioral reports. The atomic action records supply consistent variables for aggregation and comparison, while captions and reports provide descriptive context beyond the predefined categories. A tool-equipped agent model can then retrieve reports, compare action predictions across time, query clinical records, and request targeted visual analysis of selected video segments, enabling retrospective investigation (see Figure~\ref{fig:teaser}).

We evaluate \method{} using recordings and linked EHR data collected over four years at \site. We examine the accuracy of local perception, the use of these observations for longitudinal reasoning, and the clinical relevance of cohort-level behavioral predictions. On our benchmark of 397 manually annotated one-minute clips from 26 patient stays, \method{} achieves up to 84.8\% accuracy and 68.6 micro-F1 across 15 atomic action categories. To probe reasoning over patient trajectories, we introduce \mcqbenchmark{}, a clinician-informed benchmark of 144 multiple-choice questions grounded in video and clinical records from two patient stays spanning 264 hours of continuous video and EHR data.  \method{} reaches 76.9\% QA accuracy with the best proprietary LLM agent and 67.9\% QA accuracy for the best open-weight LLM agent, while no long-context VLM or Socratic caption baseline~\citep{zeng2022socratic,chandrasegaran2024hourvideo} exceeds 50\%. Finally, we apply \method{} across a cohort of 280 patients spanning approximately 60,000 camera-hours and detect significantly lower frequencies of purposeful activities among patients with a positive delirium assessment during their stay, indicating that \method{} can quantify clinically relevant behavioral differences between patient groups. %

\section{Harness Design}
\label{sec:harness}

\begin{table}[t]
  \centering
  \footnotesize
  \caption{\textbf{MCQ performance on \mcqbenchmark.} QA accuracy (\%) is shown by question type and is averaged over 5 gold-answer placements.  \method rows share an identical harness (prompts, taxonomy, and temporal hierarchy) and differ only in the agent model. Best per column is in bold, and second best is underlined. The video tool is served by the agent model, or by Gemini 3.1 Flash Lite where marked $^\dagger$. 
  }
  \label{tab:mcq-performance}
  \vspace{1mm}
  \begin{tabular}{@{}l c c c c@{}}
    \toprule
    & \cellcolor{green!18}\textbf{Video} & \cellcolor{orange!22}\textbf{EHR} & \cellcolor{blue!14}\textbf{Cross-modal} & \textbf{Avg.} \\
    \midrule
    \rowcolor{gray!12} \multicolumn{5}{@{}l}{\textit{No video}} \\
    Random & 20.1 & 20.0 & 20.0 & 20.1 \\
    Text-only LLM, Gemini 3.1 Flash Lite & 32.3 & 26.9 & 25.0 & 30.9 \\
    \rowcolor{gray!12} \multicolumn{5}{@{}l}{\textit{Long-context VLMs}} \\
    Qwen3-VL 8B, 475 frames & 31.6 & 10.0 & 47.5 & 28.6 \\
    Qwen3-VL 8B, 950 frames & 36.3 & 13.8 & 50.0 & 33.0 \\
    Gemini 3.1 Flash Lite, 950 frames & 36.5 & 14.6 & 25.0 & 31.9 \\
    Gemini 3.1 Flash Lite, 475 frames + EHR & 41.8 & 83.8 & 40.0 & 49.3 \\
    \rowcolor{gray!12} \multicolumn{5}{@{}l}{\textit{Socratic models (HourVideo)}} \\
    Qwen3-VL 8B & 33.7 & 22.3 & 20.0 & 30.9 \\
    Qwen3.5 9B & 30.4 & 18.5 & 10.0 & 27.1 \\
    Gemini 3.1 Flash Lite & 36.8 & 16.2 & 47.5 & 33.7 \\
    Gemini 3.1 Flash Lite + EHR & 37.8 & 76.2 & 42.5 & 45.0 \\
    \rowcolor{gray!12} \multicolumn{5}{@{}l}{\textit{\method harness (ours)}} \\
    \quad w/ gpt-oss-20b$^\dagger$ & 50.3 & 93.1 & 40.0 & 57.4 \\
    \quad w/ Qwen3.5 9B & 49.5 & 95.4 & \underline{52.5} & 57.9 \\
    \quad w/ Gemini 3.1 Flash Lite & 53.9 & 96.2 & 45.0 & 61.0 \\
    \quad w/ Gemma-4 31B$^\dagger$ & \underline{61.8} & \textbf{100.0} & 47.5 & \underline{67.9} \\
    \quad w/ Gemini 3.5 Flash & \textbf{72.0} & \underline{99.2} & \textbf{72.5} & \textbf{76.9} \\
    \bottomrule
  \end{tabular}
\end{table}

Following the terminology of prior work~\citep{packer2023memgpt,yang2024sweagent,lee2026metaharness}, we call the \emph{harness} everything that surrounds an LLM: the representations it can access, the tools that expose them, and the reasoning loop under which it acts.  For our setting (Figure~\ref{fig:teaser}), a harness must provide three things: representations of the stay that can capture long-term changes in patient trajectory and moment-level temporal scales, tools that expose those representations together with the raw video and clinical records, and a loop under which a model can gather evidence within a fixed budget. 

\subsection{Problem Formulation}
\label{sec:problem}

We model an ICU stay as 
$\mathcal{S}=\big(\{V^{(1)}_t, V^{(2)}_t\}_{t=1}^{T},\,\mathcal{E}\big)$, where $V^{(1)}_t$ and $V^{(2)}_t$ are the one-minute clips from two wall-mounted cameras at minute $t$, $T$ is the length of the stay in minutes, and $\mathcal{E}$ is the set of timestamped events in the electronic health record (medication administrations, bedside assessments, vital signs, etc.). In our cohort the median stay is 3.3 days ($T\approx4{,}800$) and stays run up to two weeks ($T\approx20{,}000$). Given a free-text clinical question $q$, the task is to return an answer supported by evidence that may lie anywhere in $\mathcal{S}$, in either modality, and at a temporal scale ranging from a moment in time to the entire stay. There are two primary constraints that motivate our system design.  First, a single day of video from one camera sampled at 1 FPS is approximately an order of magnitude past the context length of most existing models, which generally range from 256K to 1M tokens~\citep{bai2025qwen3,comanici2025gemini} (see Appendix~\ref{app:tokens} for more details).  Second, a model whose weights are fitted to protected health information cannot be easily shared with external institutions. Thus, we design our harness so that all problem-specific structure and performance gains live in components that can be publicly released (our activity taxonomy, timeline structure and construction, aggregation tools, and model prompts).

\begin{figure}[t]
    \centering
    \includegraphics[width=0.86\linewidth]{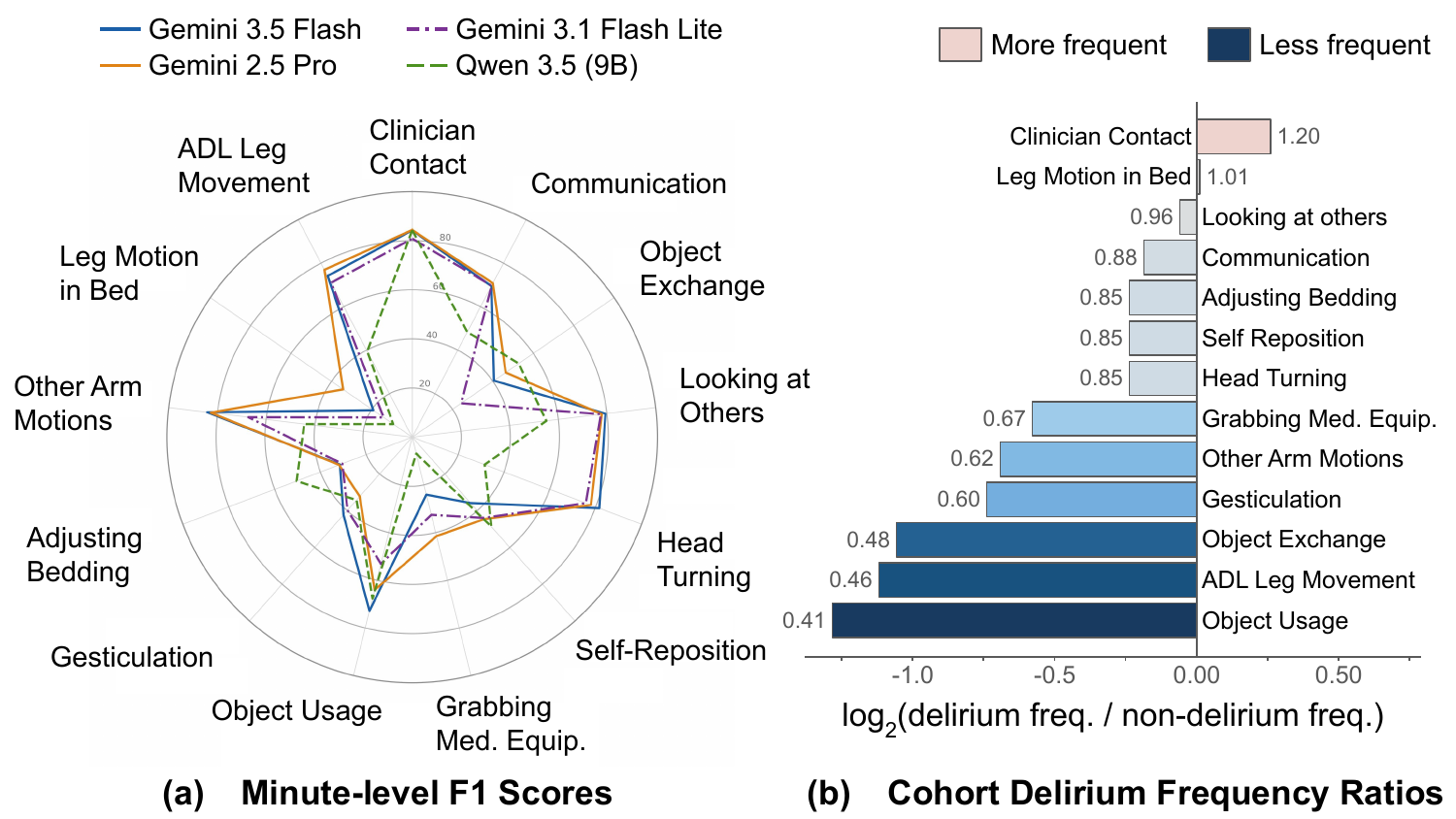}
    \caption{\textbf{Per-Category Breakdown of Atomic Action Performance.}   (a) We plot the per-category F1 scores of the atomic actions used by \method.  (b) We show the relative frequency of predicted atomic actions with respect to delirium status across a cohort of 280 patients spanning $\sim60,000$ hours of video data. Delirium is defined as any positive CAM-ICU assessment during the stay ($n=53$ delirium, $n=227$ non-delirium). Other torso movements and clinician restraining contained too few positive samples to be evaluated accurately and are therefore excluded from both panels.}
    \label{fig:atomic-action-deepdive}
\end{figure}

\method is a harness that separates \emph{temporal abstraction}, which runs continuously in the background as the stay is recorded, from \emph{evidence acquisition}, a process that is performed once per question. During temporal abstraction, as each minute of video arrives it is converted into a minute-level clinical event record consisting of an atomic-action label vector and a caption. As each hour completes, its sixty minute records are fused with the concurrent medication events into an hour-level report. During evidence acquisition, \method{} supplies a language model agent with a system prompt and a list of tools and accompanying descriptions for examining the hospital stay. Importantly, the tools operate at every level of the hierarchy, including the raw clips $V_t$ and the raw events $\mathcal{E}$. To temporally localize events, the agent uses time cues from the question, from the clinical record, from an action query over the timeline, or from reading through hourly reports (see Table~\ref{tab:time-anchor} in Appendix~\ref{app:question-creation} for an exact breakdown). The agent then chooses a tool and its arguments, receives the result in context, and repeats until it commits to an answer or exhausts a fixed budget of calls. 

\subsection{Harness Architecture}
\label{sec:architecture}

\method instantiates this harness with three model roles: a \emph{perception} VLM that turns each minute of video into a clinical event record, a \emph{report} LM that compacts each hour of records and EHR events into a report, and an \emph{agent} model that uses tools and interacts with the user. The first two run continuously as the stay is recorded to build \method{}'s temporal hierarchy, allowing the agent to gather information from individual minutes and hourly reports and to view the raw video feed and EHR. The agent model runs per question. No role is trained and each can be replaced (Appendix~\ref{app:models}).

\subsection{Minute-Level Clinical Event Records}
\label{sec:minute}

\paragraph{Taxonomy.} The structured representation of a minute is a vector over 15 \emph{atomic actions} in five groups: patient--clinician interaction (contact, talking, object exchange, restraint) and movements of the head, torso, arms, and legs (Figure~\ref{fig:teaser}). The taxonomy was developed iteratively with nurses, clinicians, and machine-learning researchers until category definitions were stable. It follows the atomic-action formulation of AVA~\citep{gu2018ava} but is chosen for clinical relevance, so that its aggregates can provide information related to mobility, agitation, or clinician attention.

\paragraph{Minute-Level Video Inference.} Each one-minute clip from each camera is sampled at 1~fps, and the resulting 60 frames are passed to the perception VLM with a fixed prompt to detect each of the 15 actions and to write a brief free-text caption of the minute (prompt and output schema in Appendix~\ref{app:minute}). The two camera views are merged by a logical OR over the 15 labels, resulting in one binary label vector of the 15 actions per patient-minute, alongside the two view-specific captions. This procedure is run over our full cohort of patients, spanning approximately 60{,}000 camera-hours, enabling the cohort-level analysis of Section~\ref{sec:results-cohort}.  The choice of perception VLM is primarily a trade-off between accuracy and cost/compute (see Appendix~\ref{app:cost}).

\subsection{Hourly Behavioral Reports}
\label{sec:hour}

As each hour of the stay completes, the report model receives its 60 minute-level clinical event records,  together with the medication administration events recorded in the EHR during that hour, and produces a structured report with six sections: (1)~a movement summary with minute-level citations, (2)~a patient--clinician interaction summary, (3)~significant moments, (4)~medication context, (5)~observable activity--medication dynamics, and (6)~a brief narrative.  We choose hour-level report generation because it matches the most frequent cadence at which entries in ICU flowsheets are typically charted. In order to test the validity of the hourly reports, we audit report faithfulness against the minute records and the medication log (see Appendix~\ref{app:faithfulness}).

\subsection{Tool-Augmented Trajectory Reasoning}
\label{sec:agent}

Given a question, the agent model runs a tool-augmented reasoning loop over the \method harness. The harness exposes four categories of tools: report, timeline, video, and EHR (exact signatures and the fixed system prompt in Appendix~\ref{app:tools}). The timeline and report tools are used to access, aggregate, or search across the temporal hierarchy, while EHR tools can access predefined categories of stored data from the EHR. The video tool enables the agent to perform additional model inference on video data to capture details that may have been missed in the atomic actions and original text caption. Specifically, the video tool allows the agent to author a prompt for a VLM that runs across both camera views (5~fps for Gemini, 2~fps for open-weight models). All output text, tool calls, and tool results are kept in the agent's context. We limit the agent to 25 tool calls, at which point a forced-choice prompt asks it to answer from the evidence gathered so far. The same prompt is used for all evaluated agent models. 

\section{Data and Evaluation}
\label{sec:eval}

\subsection{ICU Dataset}
Continuous video was recorded in the cardiovascular and medical ICUs of \site between August 2021 and June 2025 by 16 fixed, corner-mounted cameras covering eight rooms (two per room) at 1920$\times$1080 and 15~fps. The video data was linked to structured EHR data (demographics and encounter metadata, admission--discharge--transfer events, nursing flowsheet data, and the medication administration record). From 2{,}617 patients assessed for eligibility, sequential clinical and technical exclusion criteria (stay length, sedative exposure, non-neurological admission, room setup, and complete camera coverage) yielded a cohort of 280 patients spanning approximately 60{,}000 camera-hours. The full selection flow and cohort characteristics are in Appendix~\ref{app:cohort}. The study was approved by \siteirb.

\subsection{Perception Benchmark}
\begin{table}[t]
  \centering
  \caption{Average atomic action detection performance. Micro-averaged across all 15 atomic actions on the 397-clip test set. Model rows are mean $\pm$ std across 5 runs. Sampling settings for every model are listed in Appendix~\ref{app:sampling}. Best result per column in \textbf{bold}.}
  \vspace{1mm}
  \label{tab:action-detection}
  \begin{tabular}{lcccc}
    \toprule
    Model & Accuracy & F1 & Precision & Recall \\
    \midrule
    \multicolumn{5}{l}{\textit{Baselines}} \\
    Random (Match distr.) & 69.9{\tiny$\pm$0.4} & 44.8{\tiny$\pm$0.8} & 44.8{\tiny$\pm$0.8} & 44.7{\tiny$\pm$1.0} \\
    Majority Class & 77.7 & 45.2 & 68.9 & 33.6 \\
    \midrule
    \multicolumn{5}{l}{\textit{Open-source}} \\
    Qwen3-VL (8B-Instruct) & 78.0{\tiny$\pm$0.1} & 37.6{\tiny$\pm$0.4} & \textbf{84.8{\tiny$\pm$0.3}} & 24.2{\tiny$\pm$0.3} \\
    Qwen3.5 (4B) & 78.2{\tiny$\pm$0.3} & 41.0{\tiny$\pm$0.8} & 78.3{\tiny$\pm$1.7} & 27.7{\tiny$\pm$0.6} \\
    Qwen3.5 (9B) & 79.7{\tiny$\pm$0.5} & 48.4{\tiny$\pm$2.0} & 79.4{\tiny$\pm$1.0} & 34.8{\tiny$\pm$1.9} \\
    Gemma-4 (E4B-it) & 74.9{\tiny$\pm$0.1} & 22.5{\tiny$\pm$0.3} & 72.2{\tiny$\pm$1.6} & 13.4{\tiny$\pm$0.2} \\
    \midrule
    \multicolumn{5}{l}{\textit{Proprietary}} \\
    Gemini 2.5 Flash Lite & 79.0{\tiny$\pm$0.3} & 53.0{\tiny$\pm$0.7} & 68.4{\tiny$\pm$0.9} & 43.3{\tiny$\pm$0.9} \\
    Gemini 2.5 Flash & 83.2{\tiny$\pm$0.2} & 66.5{\tiny$\pm$0.5} & 71.6{\tiny$\pm$0.6} & \textbf{62.0{\tiny$\pm$0.9}} \\
    Gemini 2.5 Pro & 84.4{\tiny$\pm$0.2} & 68.1{\tiny$\pm$0.5} & 76.0{\tiny$\pm$0.4} & 61.7{\tiny$\pm$0.7} \\
    Gemini 3.1 Flash Lite & 82.3{\tiny$\pm$0.2} & 62.2{\tiny$\pm$0.3} & 74.5{\tiny$\pm$0.7} & 53.3{\tiny$\pm$0.4} \\
    Gemini 3.5 Flash & \textbf{84.8{\tiny$\pm$0.0}} & \textbf{68.6{\tiny$\pm$0.1}} & 78.3{\tiny$\pm$0.3} & 61.0{\tiny$\pm$0.3} \\
    \bottomrule
  \end{tabular}
\end{table}

To validate minute-level perception we assembled 397 one-minute clips from 26 patient stays, sampled for diversity in time of day, patient acuity, and activity level, and labeled for all 15 atomic actions by one annotator and then verified or corrected by a second ($\kappa=0.81$, 92.7\% agreement on a 20-clip audit; Appendix~\ref{app:reliability}). We report micro-averaged accuracy, F1, precision, and recall across all 15 atomic actions and 9 perception VLMs in Table \ref{tab:action-detection}.

\subsection{\mcqbenchmark: Longitudinal Reasoning Benchmark}
\label{sec:benchmark}
\paragraph{Benchmark Construction.} To evaluate reasoning over whole stays we created \mcqbenchmark, a benchmark consisting of 144 five-way multiple-choice questions grounded in 264 hours of continuous video and health records from two ICU stays. To construct the benchmark, each stay was divided into six-hour periods, and an annotator with access to the video and the health records built a timeline of events for that period. Annotators then created questions from these timelines, the record, and the raw video data in consultation with clinicians. Distractors were revised to be plausible and to defeat surface cues, and any question that blind LLMs answered correctly at least 60\% of the time was flagged for manual review and then dropped or rewritten. Each question is tagged by the modality of evidence it requires to answer correctly (\emph{video}, \emph{EHR}, and \emph{cross-modal}). During evaluation we report the average scores across 5 runs, with each run using a different position for the ground truth answer. All reported results use a single fixed set of prompts and tools. The agent prompt development procedure is described in Appendix~\ref{app:question-creation}. The report prompt, tool set, atomic action taxonomy, and temporal hierarchy were designed without reference to either benchmark. Example questions are in Appendix~\ref{app:questions}.  

  \paragraph{Baselines.} For the \emph{text-only LLM} baseline, the agent has no access to any data and attempts to answer the question solely based on language priors. \emph{Long-context VLMs} receive the question together with 475 or 950 frames sampled uniformly across the entire stay (Qwen3-VL 8B; Gemini 3.1 Flash Lite). \emph{Socratic Models}~\citep{zeng2022socratic} baselines follow the protocol proposed in HourVideo~\citep{chandrasegaran2024hourvideo}, where a VLM captions every minute of the stay and those captions are used by a language model to answer a question. Captions are temporally subsampled if their aggregate exceeds the LLM context length. We also show results for both long-context and Socratic baselines when including the stay's EHR in context (ranging from 130k-420k tokens).  We note that these baselines use roughly three times more tokens than \method{} during question answering  while achieving substantially lower accuracies. We include a more in depth discussion about baselines and compute budget in Appendix~\ref{app:socratic-ehr}. Baselines use their own fixed prompts, and additional implementation details for each baseline are included in Appendix~\ref{app:baselines}. 

\section{Experiments and Results}
\label{sec:results}

In order to evaluate \method as a harness, we perform experiments at three levels of granularity. First, we describe the experiments and results for minute-level perception evaluation in Section \ref{sec:results-minute-level}. Section \ref{sec:results-qa} describes the experiments and results on our \mcqbenchmark long-form MCQ benchmark. Finally, we report our findings when applying \method across our entire 280 patient cohort in Section \ref{sec:results-cohort}.

\subsection{Minute-level Perception}
\label{sec:results-minute-level}

Table~\ref{tab:action-detection} summarizes aggregate performance on our atomic action benchmark spanning 397 one-minute clips. Gemini 3.5 Flash performs best among the evaluated models, achieving 84.8\% accuracy and a micro-F1 score of 68.6. Gemini 3.1 Flash Lite achieves 82.3\% accuracy and a micro-F1 score of 62.2 at less than one-tenth the cost, motivating its usage as the perception VLM when applying \method{} across our 280 patient cohort. We include more details about the cost-performance tradeoff in Appendix \ref{app:cost}. We show per-category performance of selected models in Figure~\ref{fig:atomic-action-deepdive}(a). 

\subsection{Longitudinal Reasoning on \mcqbenchmark}
\label{sec:results-qa}

\paragraph{Comparison with baselines.}  We note the performance of Gemini 3.1 Flash Lite only reaches 45.0\% using the Socratic models approach, only 49.3\% when using raw video frames and EHR, and climbs to 61.0\% when using \method.  Baselines without the EHR score within 2.8 points of the text-only LLM. Although access to EHR raises the overall accuracy of baselines, they still score at most 83.8\% on EHR-related questions, while \method{} scores 96.2\%. With the same language model and the same data sources, \method{}'s hierarchy and tool use outperform the long-context and Socratic baselines (+11.7 points over the strongest baseline; 95\% CI [4.4, 18.9], paired bootstrap over questions).

\begin{table}[t]
  \centering
  \footnotesize
  \caption{\textbf{Harness ablation.} Accuracy (\%) averaged over three agent models: Gemini 3.5 Flash, Gemini 3.1 Flash Lite, and Qwen3.5 9B. Single-toolset rows give the agent one tool family; cumulative rows add one family at a time up to the full harness, with gains shown in green. Best per column and block in bold, second best underlined. Additional results in Appendix~\ref{app:ablation-models} and~\ref{app:toolsalone}.}
  \label{tab:mcq-ablation}
  \vspace{1mm}
  \begin{tabular}{l ccc c c c >{\scriptsize\color{ForestGreen}}c}
    \toprule
    & \multicolumn{3}{c}{\cellcolor{green!18}\textbf{Video}} & \cellcolor{orange!22}\textbf{EHR} & \cellcolor{blue!14}\textbf{Cross-modal} & \multicolumn{1}{c}{\textbf{Avg.}} & \multicolumn{1}{c}{} \\
    Agent tools & $<$5 min & 5--30 min & $>$30 min & Record only & Video + EHR & & \\
    \midrule
    \rowcolor{gray!12} \multicolumn{8}{l}{\textit{Single toolset performance}} \\
    Timeline only & 27.5 & 43.3 & \textbf{60.7} & 30.3 & 25.0 & \underline{43.7} &  \\
    Reports only & \underline{33.7} & \textbf{52.5} & \underline{47.4} & \underline{33.3} & \underline{43.3} & 42.1 &  \\
    EHR only & 32.9 & 31.7 & 41.2 & \textbf{96.4} & \textbf{60.0} & \textbf{49.3} &  \\
    Video only & \textbf{39.8} & \underline{50.8} & 38.3 & 25.1 & 27.5 & 37.0 &  \\
    \midrule
    \rowcolor{gray!12} \multicolumn{7}{l}{\textit{Cumulative toolset performance}} & \multicolumn{1}{c}{\scriptsize$\Delta$} \\
    w/ timeline & 27.5 & 43.3 & 60.7 & 30.3 & 25.0 & 43.7 & 0.0 \\
    + reports & 30.5 & \underline{55.4} & 63.8 & 31.3 & 31.7 & 47.6 & +3.9 \\
    + EHR & \underline{31.7} & 51.6 & \underline{64.1} & \textbf{97.2} & \textbf{61.7} & \underline{61.1} & +17.4 \\
    \rowcolor{blue!8} + video (\method) & \textbf{39.2} & \textbf{57.1} & \textbf{69.2} & \underline{96.9} & \underline{56.7} & \textbf{65.3} & +21.6 \\
    \bottomrule
  \end{tabular}
\end{table}

\paragraph{Harness ablation.} We ablate the harness across three different agent models under two settings (Table~\ref{tab:mcq-ablation}): single toolset and cumulative. In our single toolset ablation, we demonstrate that each toolset contributes most to performance for specific types of questions (Table~\ref{tab:mcq-ablation}, top). EHR tools significantly improve the performance on EHR-related questions, video tools allow for greater understanding of short moments in time spanning $<$5 minutes, reports allow for the agent to more accurately answer questions spanning 5 to 30 minutes in length, and the timeline tools are most helpful for questions that require reasoning across longer time horizons. Adding toolsets cumulatively (Table~\ref{tab:mcq-ablation}, bottom) yields sustained gains in average accuracy with each toolset added, and the full harness exceeds the best single toolset by 16 points. The best performance achieved by \method is using Gemini 3.5 Flash where it reaches up to 76.9\% on \mcqbenchmark. Additional per-agent results, paired tests, and per-stay breakdowns are in Appendices~\ref{app:ablation-models} and~\ref{app:toolsalone}.

\paragraph{Agentic performance.}
To isolate the contributions of the agent LLM from the perception layer and temporal abstraction process, we conduct an experiment using Gemini 3.1 Flash Lite for video perception and temporal abstraction (minute-level record generation, hour reports, and all video-tool queries) while varying the agent LLM. Under this setting, open-weight models including Qwen3.5 9B and Gemma-4 31B match or outperform Gemini 3.1 Flash Lite as the agent LLM (Table~\ref{tab:fixed-perceiver}). Together with the action detection results in Table~\ref{tab:action-detection}, our findings indicate that the performance gap for evaluated open-weight models exists primarily from limitations in perception rather than agentic reasoning.

\begin{table}[t]
  \centering
  \footnotesize
  \caption{\textbf{Comparison across agent models with fixed perception capabilities.} Accuracy (\%) on \mcqbenchmark with Gemini 3.1 Flash Lite used for all video tool calls, timeline,  and report construction. Best per column in bold, second best underlined.}
  \label{tab:fixed-perceiver}
  \vspace{1mm}
  \begin{tabular}{l ccc c c c c}
    \toprule
    & \multicolumn{3}{c}{\cellcolor{green!18}\textbf{Video}} & \cellcolor{orange!22}\textbf{EHR} & \cellcolor{blue!14}\textbf{Cross-modal} & \textbf{Avg.} & \textbf{Calls} \\
    Agent model & $<$5 min & 5--30 min & $>$30 min & Record only & Video + EHR & & \\
    \midrule
    \rowcolor{gray!12} \multicolumn{8}{l}{\textit{Proprietary}} \\
    Gemini 3.5 Flash & 35.8 & \underline{57.5} & \textbf{80.3} & \underline{98.5} & \textbf{77.5} & \textbf{70.7} & 9.5 \\
    Gemini 3.1 Flash Lite & \underline{36.4} & 52.5 & 63.7 & 96.2 & 45.0 & 61.0 & 8.9 \\    \rowcolor{gray!12} \multicolumn{8}{l}{\textit{Open-weight}} \\
    gpt-oss-20b & 32.1 & 41.2 & 62.5 & 93.1 & 40.0 & 57.4 & 3.7 \\
    Qwen3.5 9B & 35.2 & 45.0 & 67.3 & 95.4 & \underline{50.0} & 61.6 & 9.0 \\    Gemma-4 31B & \textbf{40.0} & \textbf{61.2} & \underline{73.7} & \textbf{100.0} & 47.5 & \underline{67.9} & 5.4 \\
    \bottomrule
  \end{tabular}
\end{table}

\subsection{Cohort-Level Behavioral Signal}
\label{sec:results-cohort}

To determine whether \method can surface clinical signal at scale, we run its minute-level perception over the full cohort of 280 patients (roughly 60{,}000 camera-hours) and compare the frequency of each atomic action between patients with at least one positive delirium assessment during their stay ($n=53$) and those with none ($n=227$). We investigate only the 13 actions with sufficient positive samples in the perception benchmark, excluding other torso movements and clinician restraining. For each action we report the \emph{delirium frequency ratio}, the per-minute frequency in the delirium group divided by that in the non-delirium group (see Appendix~\ref{app:delirium-analysis} for more details). We visualize each evaluated atomic action's delirium frequency ratio in Figure~\ref{fig:atomic-action-deepdive}(b). Seven actions are significantly less frequent among patients with documented delirium, and all but head turning remain significant after Holm correction on per-patient rates (Appendix~\ref{app:delirium-analysis}). The largest reductions are in purposeful, goal-directed actions: using non-medical objects ($0.41\times$), leg movements associated with daily living ($0.46\times$), object exchange with a clinician ($0.48\times$), and gesticulation ($0.60\times$). The fraction of observed minutes containing any of these four actions is lower in the delirium group (median 0.053 vs.\ 0.146 per patient, Mann--Whitney $p = 1.3\times10^{-8}$). Clinician--patient contact is the only atomic action that is more frequent for the delirium group ($1.20\times$), consistent with greater bedside attention. We include patient-level bootstrap intervals and an equal-patient-weight sensitivity analysis in Appendix \ref{app:delirium-analysis}.

\section{Related Work}
\label{sec:related}

\paragraph{Ambient intelligence in hospitals.} Computer vision has been applied to inpatient monitoring through depth sensors and accelerometry for ICU mobility~\citep{ma2017measuring,mehra2017depth}, supervised detection of patient mobilization~\citep{yeung2019computer}, pervasive multimodal sensing~\citep{davoudi2019intelligent}, a supervised behavioral atlas of 40 ICU activities~\citep{dai2025developing}, and commercial real-time video analysis on general wards~\citep{gabriel2025continuous}. These works exist within the broader research direction of ambient intelligence for healthcare~\citep{haque2020illuminating,yeung2018bedside}. In general, these systems train task-specific models on curated annotations using real patient data for a fixed set of behaviors, so the resulting models are fit to protected health information and are difficult to share publicly. In comparison with prior work, \method proposes a shareable harness for pretrained models and fuses the activity detections with the EHR to support question-answering over entire stays.

\paragraph{Long-video understanding.} Video understanding benchmarks have recently extended from minutes~\citep{mangalam2023egoschema} to an hour or more in length~\citep{fu2024videomme,wu2024longvideobench,chandrasegaran2024hourvideo}. Existing long-form video understanding systems are generally designed for that regime and either compose frozen models through language~\citep{zeng2022socratic,zhang2023simple,wang2025videotree}, let an agent retrieve and re-inspect segments~\citep{wang2024videoagent,fan2025videoagent,zhang2025deepvideodiscovery}, or maintain a multi-level caption memory with backtracking~\citep{zuo2025videolucy}. Week-long egocentric life-logs are the closest precedent to our work: EgoLife summarizes captions at hourly and daily levels and retrieves from day to hour to clip, with a fine-tuned captioner~\citep{yang2025egolife}, while Ego-R1 trains an agent by reinforcement learning to navigate a week/day/hour hierarchy~\citep{tian2025egor1}. \method differs from existing works in both our setting and by fusing a second asynchronous modality (EHR) into our temporal  hierarchy.

\paragraph{Clinical agents.} LLM agents in medicine generally operate over text or structured data. Existing works focus on multi-agent question answering~\citep{tang2024medagents}, code generation over electronic records~\citep{shi2024ehragent}, tool orchestration over single images~\citep{fallahpour2025medrax}, and diagnostic dialogue~\citep{tu2025towards}. Recent work finds that LLM agents struggle to understand ICU stays from the EHR alone ~\citep{shen2026realicu}. \method extends upon prior work by examining the complementary information provided by days of continuous bedside video alongside EHR, rather than viewing each modality in isolation.

\section{Discussion}
\label{sec:discussion}
\paragraph{Key findings.} VLMs detect ICU atomic actions zero-shot with useful, if somewhat noisy, accuracy (68.6 F1). We demonstrate that these predictions, when combined with the EHR, can be used to construct a training-free LLM harness for greater understanding of patient stays. Specifically, \method enables LLM agents to more accurately answer questions over multi-day stays where existing approaches struggle (61.0\% vs.\ at most 49.3\% for the same backbone). When ablating our harness, we find that each toolset is best for different question archetypes. When run over the full 280-patient cohort, we show that the ICU atomic action detections provide signal for separating patients by delirium status. Purposeful actions such as using objects with hands and gesticulation are significantly less frequent among patients with documented delirium. Finally, we find that the gap between open-weight and proprietary models lies primarily in perception. When the perception layer is fixed, open-weight agents can match or even exceed the performance of proprietary models (Table~\ref{tab:fixed-perceiver}). This has significant implications for on-premise deployment, which we find is currently limited more by the performance of the perception VLM than by the agent model.

\paragraph{Behavioral signal and delirium.} The cohort-level associations shown in this work are consistent with the motoric phenomenology of delirium: patients with a documented positive screen show fewer purposeful, goal-directed actions. Hypoactive delirium, defined by reduced psychomotor activity, is both the most prevalent subtype in the ICU~\citep{krewulak2018incidence} and the one most often missed by intermittent screening. Furthermore, the  sensitivity for its detection has been shown to degrade in routine practice~\citep{vaneijk2011routine}. We emphasize that these are associations computed from model predictions on retrospective data; they are hypothesis-generating and motivate future work but do not by themselves establish predictive capabilities. 

\paragraph{User Interface.}  Because ICUs already suffer from alarm fatigue~\citep{sendelbach2013alarm}, \method is designed to give clinicians new capabilities for retrospective querying rather than serving as an engine for continuous alerting. The harness was built to run inside a viewer for bedside video data, which displays the detected atomic actions as a timeline underneath the video and lets clinicians  ask questions in a chat panel. Clinicians can see the agent's tool calls and can click any cited timestamp to skip to the corresponding minute of video. We have demonstrated these capabilities to ICU nurses and physicians and used their feedback to guide the interface design. 

\paragraph{Limitations.} This is a single-center retrospective study; camera placement, population, and documentation practice may not transfer. \mcqbenchmark is deep rather than broad: 144 questions over two different ICU stays. Delirium labels come from intermittent bedside tests, and per-action detection error propagates into the detected delirium frequency ratios. Hourly reports are model-generated and may contain inaccuracies, hallucinations, or mistakes (see Appendix~\ref{app:faithfulness}). Continuous monitoring raises consent, staff-surveillance, and fairness questions~\citep{haque2020illuminating,martinezmartin2021ethical} that we did not explicitly discuss in this work but are prerequisites for deployment.

\paragraph{Reproducibility.} The harness, tools, prompts, and taxonomy are released at \url{https://github.com/zanedurante/tracer-harness}. Because we do not train any part of our harness on protected health data, the same hierarchy can be instantiated over any other temporally indexed video or health record collection.  We include instructions for how to run our harness on sample videos and data in the linked repository.

\ifpreprint\else\newpage\fi
\section*{AI use statement}
Large language model assistants (ChatGPT, Claude, and Claude Code) were used in preparing this work, in accordance with the ICLR AI Policy for Authors. \textit{Writing:} AI assistants drafted and revised parts of the manuscript, edited text for readability and length, suggested structure for sections, converted the manuscript between formatting templates, formatted and verified bibliography entries, and checked consistency of terminology, numbers, and cross-references. \textit{Code and experiments:} AI assistants helped implement the methods and baselines described in the paper, data cleaning and reformatting, debugging, and analysis code. The research questions, harness design, and atomic action taxonomy were developed by the authors and clinical collaborators, and all annotations were produced manually. \mcqbenchmark questions were written by annotators in consultation with clinicians from manually constructed event timelines; AI assistants were used to help draft and revise the wording of some questions and distractors, but every question was reviewed and verified by the authors. Every AI-assisted passage in the paper was reviewed and edited by the authors, who take full responsibility for the content. 
\ifpreprint
\section*{Acknowledgments}
We thank the clinical staff of the cardiovascular and medical ICUs at Stanford Hospital and the hospital engineering teams who installed and maintained the bedside sensing infrastructure. We thank Rachel Karp, Lauren (Lo) Wesson, Julie Glover, Patricia (Trish) Suits, and Heather McKee Hickman for their insights and feedback that shaped the direction of this project. This research was partially supported by a research award from Schmidt Futures (research support), the Stanford Center for Digital Health (research support), Stanford Medicine (support for building the infrastructure and camera sensors), and Google Cloud credits from the Stanford Institute for Human-Centered Artificial Intelligence (HAI) Google Cloud Credit Grants program. The authors acknowledge the National Artificial Intelligence Research Resource (NAIRR) Pilot (award NAIRR240024) for providing compute resources for this project. This research used resources of the Oak Ridge Leadership Computing Facility, which is a DOE Office of Science User Facility supported under Contract DE-AC05-00OR22725 (project CSC659\_MDE).
\fi
\newpage

\bibliography{bibliography}
\bibliographystyle{iclr2027_conference}

\clearpage
\appendix
\appendix

\section{Appendix}

\subsection{\mcqbenchmark{} Example Questions}
\label{app:questions}
\label{app:examples}
We include the following de-identified questions from \mcqbenchmark{}. We show two questions per archetype. Calendar dates are replaced by the day of the stay and clock times by [time]. 

\begin{mcqcard}{vidd}{Video moment ($<$5 min) \quad stay B}
At [time] on day 3 of the stay, what clinical object was removed from the patient?
\begin{enumerate}[label=(\Alph*), leftmargin=1.8em, itemsep=0pt, topsep=2pt, parsep=0pt]
  \item An SCD sleeve
  \item An endotracheal tube
  \item The ventilator (extubation)
  \item A heel protector boot \correct
  \item A central line
\end{enumerate}
\end{mcqcard}

\begin{mcqcard}{vidd}{Video moment ($<$5 min) \quad stay B}
From the observable video, around [time] on day 1 of the stay, what kind of bedside imaging was performed?
\begin{enumerate}[label=(\Alph*), leftmargin=1.8em, itemsep=0pt, topsep=2pt, parsep=0pt]
  \item An MRI
  \item A PET scan
  \item An ultrasound scan \correct
  \item A CT scan
  \item An X-ray
\end{enumerate}
\end{mcqcard}

\begin{mcqcard}{vidd}{Video window (5--30 min) \quad stay B}
Over the 10 minutes from about [time] on day 2 of the stay, the patient was predominantly:
\begin{enumerate}[label=(\Alph*), leftmargin=1.8em, itemsep=0pt, topsep=2pt, parsep=0pt]
  \item Being handled by staff at the bedside (repositioning, linen or line adjustments), with only slight movements of her own
  \item Completely still, with no head, eye, or limb movement at all \correct
  \item Awake and talking with staff, gesturing with her hands
  \item Still except for occasional small hand or finger movements
  \item Repeatedly moving her arms and head on her own with no staff at the bedside
\end{enumerate}
\end{mcqcard}

\begin{mcqcard}{vidd}{Video window (5--30 min) \quad stay A}
Between about [time] and [time] on day 2 of the stay, no staff were in the room. Which of these most accurately describes the patient's behavior?
\begin{enumerate}[label=(\Alph*), leftmargin=1.8em, itemsep=0pt, topsep=2pt, parsep=0pt]
  \item Repeatedly pulling at the tubing at her face until a nurse came in
  \item Talking with a visitor seated at the bedside
  \item Awake and using a smartphone while lying in bed \correct
  \item Sitting up on the edge of the bed
  \item Asleep with her eyes closed and her hands under the blanket
\end{enumerate}
\end{mcqcard}

\begin{mcqcard}{vidd}{Long-form video ($>$30 min) \quad stay B}
Which of the following sequences correctly describes the order of events during the ICU recording?
\begin{enumerate}[label=(\Alph*), leftmargin=1.8em, itemsep=0pt, topsep=2pt, parsep=0pt]
  \item Patient taken out for tests and later returned $\rightarrow$ Ultrasound/scanning performed $\rightarrow$ Family members enter the room $\rightarrow$ SCD compression sleeves placed on the legs
  \item Ultrasound/scanning performed $\rightarrow$ SCD compression sleeves placed on the legs $\rightarrow$ Family members enter the room $\rightarrow$ Patient taken out for tests and later returned
  \item Ultrasound/scanning performed $\rightarrow$ Family members enter the room $\rightarrow$ Patient taken out for tests and later returned $\rightarrow$ SCD compression sleeves placed on the legs \correct
  \item SCD compression sleeves placed on the legs $\rightarrow$ Ultrasound/scanning performed $\rightarrow$ Family members enter the room $\rightarrow$ Patient taken out for tests and later returned
  \item Family members enter the room $\rightarrow$ Ultrasound/scanning performed $\rightarrow$ Patient taken out for tests and later returned $\rightarrow$ SCD compression sleeves placed on the legs
\end{enumerate}
\end{mcqcard}

\begin{mcqcard}{vidd}{Long-form video ($>$30 min) \quad stay B}
How did the patient's level of voluntary movement and interaction before being taken out for tests compare to after returning to the room?
\begin{enumerate}[label=(\Alph*), leftmargin=1.8em, itemsep=0pt, topsep=2pt, parsep=0pt]
  \item The patient did not visibly interact before the tests but later spoke with clinicians after returning to the room.
  \item The patient appeared active and responsive both before and after the tests.
  \item The patient was mostly motionless before the tests but later sat up in a slanted position in the bed after returning to the room.
  \item The patient remained largely motionless both before and after the tests, with no clear voluntary movement observed. \correct
  \item The patient spoke with family members before being taken for tests and later spoke with clinicians after returning to the room.
\end{enumerate}
\end{mcqcard}

\begin{mcqcard}{ehrd}{EHR \quad stay A}
Besides sedatives and pain medication, the patient received a class of scheduled medications throughout the stay. Which class?
\begin{enumerate}[label=(\Alph*), leftmargin=1.8em, itemsep=0pt, topsep=2pt, parsep=0pt]
  \item Antibiotics
  \item Blood thinners (anticoagulants)
  \item Corticosteroids (steroids)
  \item Insulin
  \item Anti-seizure (antiepileptic) medications \correct
\end{enumerate}
\end{mcqcard}

\begin{mcqcard}{ehrd}{EHR \quad stay A}
The first CAM-ICU delirium assessment, on the first night of the stay, returned what result?
\begin{enumerate}[label=(\Alph*), leftmargin=1.8em, itemsep=0pt, topsep=2pt, parsep=0pt]
  \item Positive (delirium detected)
  \item Negative (no delirium detected)
  \item Unable to assess (UTA) \correct
  \item Negative on the initial screen, then re-assessed as positive later the same shift
  \item The assessment was attempted but the patient's RASS was too low ($-4$ or $-5$) to proceed
\end{enumerate}
\end{mcqcard}

\begin{mcqcard}{crossd}{Cross-modal \quad stay A}
During the evening of day 2 of the stay, there was a stretch of about ten minutes with no staff in the room during which the patient used a smartphone while lying in bed. At about what time was that stretch, and which medication was administered within 45 minutes after it ended?
\begin{enumerate}[label=(\Alph*), leftmargin=1.8em, itemsep=0pt, topsep=2pt, parsep=0pt]
  \item {}[time window]; the norepinephrine infusion was stopped afterwards
  \item {}[time window]; no medication followed within 45 minutes
  \item {}[time window]; an albuterol--ipratropium nebulizer treatment followed \correct
  \item {}[time window]; a norepinephrine rate change followed
  \item {}[time window]; a chlorhexidine oral rinse followed
\end{enumerate}
\end{mcqcard}

\begin{mcqcard}{crossd}{Cross-modal \quad stay B}
During the morning of day 2 of the stay, when did the patient have her period(s) of wakefulness?
\begin{enumerate}[label=(\Alph*), leftmargin=1.8em, itemsep=0pt, topsep=2pt, parsep=0pt]
  \item Multiple short awakenings starting around [time], each lasting only a few seconds
  \item Briefly around [time] and again around [time], with sedation in between \correct
  \item Around [time] only, with no other alert periods that morning
  \item A single stretch from about [time] through late morning
  \item Briefly around [time] only, then remaining sedated through the afternoon
\end{enumerate}
\end{mcqcard}

\clearpage
\subsection{Tool Interface and Agent Prompt}
\label{app:tools}
\paragraph{Tools.} Table~\ref{tab:tools} lists every tool with its exact name, parameters (a question mark marks an optional parameter; \texttt{pat\_id} defaults to the current patient and is omitted from the table except where it is the only parameter), and return value. We group tools by the families used in the ablations. Each tool configuration ablated in Table~\ref{tab:mcq-ablation} consists of the scaffold tool set plus the families listed. Times are given in the site's local time zone (``local'' in tool names and prompt text; the released code sets the zone through \texttt{TRACER\_LOCAL\_TZ}) and hours are counted from ICU admission.

\begin{table}[h]
  \centering
  \footnotesize
  \caption{\textbf{The agent's tools.} Exact names, parameters, and return values of the eleven tools, grouped by family; the row labels of Table~\ref{tab:mcq-ablation} name the families each configuration adds to the scaffold pair. A question mark marks an optional parameter.}
  \label{tab:tools}
  \vspace{1mm}
  \renewcommand{\arraystretch}{1.12}
  \begin{tabularx}{\linewidth}{@{} l >{\raggedright\arraybackslash\hsize=0.9\hsize}X >{\raggedright\arraybackslash\hsize=1.1\hsize}X @{}}
    \toprule
    Tool & Parameters & Returns \\
    \midrule
    \multicolumn{3}{@{}l}{\textbf{Scaffold} (every configuration)} \\
    \texttt{get\_patient\_metadata} & \texttt{pat\_id}? & demographics, ICU admission time, observation window, encounter history \\
    \texttt{convert\_utc\_local} & \texttt{dt\_str}, \texttt{to\_timezone} & the datetime converted between UTC and local time \\
    \addlinespace[3pt]
    \multicolumn{3}{@{}l}{\textbf{Timeline} (minute-level records)} \\
    \texttt{get\_atomic\_actions} & \texttt{hour\_index} & per-minute active action labels for one hour from admission \\
    \texttt{query\_actions} & \texttt{labels}?, \texttt{start\_hour}?, \texttt{end\_hour}? & minutes active per label over the range; per-hour counts; per-label per-hour minutes \\
    \texttt{find\_first\_occurrence} & \texttt{labels}, \texttt{start\_hour}?, \texttt{end\_hour}? & first minute where any listed label is active (hour, minute, timestamp) \\
    \texttt{find\_moments\_of\_interest} & \texttt{target\_labels}?, \texttt{start\_pt}?, \texttt{end\_pt}?, \texttt{min\_rise\_pct}?, \texttt{drug\_window\_min}? & rises and falls of a 15-minute rolling activity signal, with the medications given nearby \\
    \addlinespace[3pt]
    \multicolumn{3}{@{}l}{\textbf{Report} (hourly reports)} \\
    \texttt{list\_hourly\_reports} & \texttt{pat\_id}? & the (date, hour) pairs with a report \\
    \texttt{get\_hourly\_report} & \texttt{date}, \texttt{hour} & the six-section report for that hour \\
    \addlinespace[3pt]
    \multicolumn{3}{@{}l}{\textbf{EHR}} \\
    \texttt{query\_medications} & \texttt{start\_pt}, \texttt{end\_pt}, \texttt{name\_filter}?, \texttt{class\_filter}?, \texttt{max\_rows}? & medication administration events in the window: time, name, action, dose, route, class \\
    \texttt{query\_flowsheet} & \texttt{disp\_name\_filter}?, \texttt{start\_pt}?, \texttt{end\_pt}?, \texttt{max\_rows}? & flowsheet rows (time, measurement, value, unit); without a filter, the measurement names available \\
    \addlinespace[3pt]
    \multicolumn{3}{@{}l}{\textbf{Video}} \\
    \texttt{query\_video\_minute} & \texttt{minute\_pt}, \texttt{prompt} & the perception VLM's answer to the prompt for each of the two camera views of that minute \\
    \bottomrule
  \end{tabularx}
\end{table}

\paragraph{Agent prompts.} The box below is the agent system prompt for the \method{} harness. Reduced tool configurations receive the same prompt minus the lines of the tools they do not expose and minus any working rule that names an absent tool. The text-only baseline receives the short prompt shown afterwards. The same prompt is used for all results across every agent model in this work. During MCQ evaluation, the agent receives the question as the user message shown underneath the agent and blind-LLM prompts.  Finally, we show the prompt for when a question exhausts the 25-call budget and a forced-choice message is shown to the agent. The instructions to always choose an option exist only for multiple-choice scoring, and in the clinical interface the agent may report that the evidence is insufficient.

\begin{promptbox}[colback=gray!6]{Agent system prompt and tool descriptions, full harness (placeholders for patient metadata)}
\begin{lstlisting}[style=prompt]
You are an analytic assistant embedded in a clinical video-viewer app for ICU patients. Answer the multiple-choice question about the patient currently on screen. Investigate with your tools BEFORE deciding; base your answer on what the tools actually return, not on priors.

Tools available to you:
- get_patient_metadata: Get demographics, ICU admission time, observation window length, and encounter history for a patient. Defaults to the patient currently on screen if pat_id is omitted.
- convert_utc_local: Convert between UTC and local time.
- list_hourly_reports: List all hourly markdown reports available for a patient. Each entry has a date (YYYY-MM-DD) and an hour (0-23). Defaults to the current patient if pat_id is omitted.
- get_hourly_report: Fetch the full markdown content of a single hourly report. Reports follow a fixed 6-section structure: Patient Movement Summary, Patient-Clinician Interaction, Significant Moments, Medication Context, Activity-Medication Alignment, Brief Summary. Defaults pat_id to the current patient.
- get_atomic_actions: Return minute-by-minute atomic action detections for one hour of the patient's stay. hour_index=0 is the first hour from ICU admission. Each minute lists currently-active action labels (from the atomic action taxonomy). Defaults pat_id to the current patient.
- query_actions: Aggregate atomic action counts across a range of hours. Returns: (a) total minutes each label was active over the window, (b) per-hour counts of minutes with any selected label active, and (c) per-label per-hour minutes active (dict[label] -> list over the hour range) -- use (c) to answer questions like 'which hour had the most X'. Filter by specific labels or omit to include all. start_hour/end_hour are zero-indexed from ICU start (end_hour is exclusive). Defaults pat_id to the current patient.
- find_first_occurrence: Find the earliest minute where any of the given action labels was active. Returns hour_index, minute (0-59), and UTC timestamp. Prefer this over reading hourly reports or scanning atomic actions when the user asks 'when did X first happen'. Optionally restrict to a range via start_hour/end_hour (zero-indexed from ICU start, end_hour exclusive). Defaults pat_id to the current patient.
- find_moments_of_interest: Detect 'moments of interest' -- rises and falls in a 15-minute rolling signal built from the atomic-action labels you select -- and cross-reference each detected blurp with medication events given nearby. The signal at minute t is the fraction of the previous 15 minutes in which any selected label fired. **You must choose target_labels to match the question.** Examples:
- query_medications: Return medication-administration-record (MAR) events for the current patient in a local-time window. Each event has taken_time_pt, med_name (display name), mar_action (Given / New Bag / Bolus / Stopped / etc.), dose (rate + unit or bolus + unit), route, and drug class (pharmacologic + therapeutic). Use this to answer questions like 'which meds were administered +/-45 min of X?', 'when was the last dose of oxycodone?', or 'what sedatives were running at 03:00?'. TIME: pass local-time strings 'YYYY-MM-DD HH:MM' (24-hour). Windows are inclusive on both ends. OPTIONAL FILTERS: name_filter (case-insensitive substring on the med name) and class_filter (substring on pharmacologic/therapeutic class, e.g. 'OPIOID', 'SEDATI', 'ANTIBIOTIC'). Results are capped at max_rows (default 50).
- query_flowsheet: Query nursing-flowsheet measurements for the current patient: vitals (Heart Rate, Pulse, SpO2, Resp, Arterial Systolic BP, Mean Arterial Pressure, Arterial Diastolic BP, BP), sedation/agitation (Patient RASS Score, Anxiety Level, Pain Scale Type), and nursing assessments (Activity, ADLs, Communication, Ectopy, Cardiac Rhythm, etc.). Each row has recorded_time_pt, disp_name, value, and unit. DISCOVERY: call without disp_name_filter to see which measurement types exist for this patient (list of disp_name + count). QUERY: pass disp_name_filter (case-insensitive substring, e.g. 'RASS', 'Heart Rate', 'SpO2') to fetch the actual measurements. Optionally scope with start_pt / end_pt (local-time 'YYYY-MM-DD HH:MM'). Use this for questions like 'when was the patient most deeply sedated?' (RASS), 'what was the peak heart rate?', 'when did BP swings occur?', 'what was the anxiety/pain trend?'.
- query_video_minute: Look at the ICU surveillance video for ONE specific minute and answer a free-form question about what is visible -- patient movements/reactions, what a clinician is doing, devices/objects, body position, who is in the room.
  Treat its answer as EVIDENCE TO WEIGH, not ground truth: the video reader can miss brief events or misread motion, and the two camera views may disagree. When the answer matters, re-query (a different prompt or an adjacent minute) and cross-check against your other tools before committing.
  This tool is useful for zooming in on a single minute of interest. You can call it on the SAME minute several times with DIFFERENT prompts to probe different details (e.g. first ask what the clinician is doing, then whether the patient moved). It returns BOTH camera angles (left and right view) answered separately.
  TIME: everything here is local time. Pass the minute exactly as it appears in the question, 'YYYY-MM-DD HH:MM' (24-hour, local time); the tool finds the right clip for you and you never convert timezones. The clip is one minute long, so if an event might straddle a boundary, also query the adjacent minute.

If a tool returns an error or nothing useful, adjust and try another approach. Tools that take a patient id default to the current patient -- omit pat_id. You MUST select exactly one option as your single best guess, even if you are unsure or the data is incomplete. Never refuse and never say you cannot determine the answer.

Patient context (from get_patient_metadata):
- Video and activity data (hourly reports, atomic actions, query_video_minute) cover {start} to {end} local time ({n_hours} hours).
- hour_index counts whole hours from {start}: {hour_examples}. To convert a time to hour_index, subtract {start} and take the whole number of hours; write the subtraction down.
- EHR data (query_medications, query_flowsheet) covers the whole hospital admission, {ehr_start} to {ehr_end} local time, which extends beyond the video window.
Every date or time in a question refers to THIS stay in local time. Never guess a year or a date outside these windows.

Working rules (follow them strictly):
1. For "during the stay", "first", "last", "how many", "which day" questions about medications or flowsheet data, query the ENTIRE admission in ONE call (start_pt = {ehr_start}, end_pt = {ehr_end}); for activity data use start_hour 0 to end_hour {n_hours}. Do not step through the stay day by day.
2. Zero rows or zero events never prove absence. Before concluding "none", re-check the dates against the patient context, remove filters, widen the window, and try a shorter substring. If a result carries a "hint" field, follow it.
3. Medications: use name_filter ALONE with a short substring of the drug name (e.g. 'norepi', 'oxycod', 'hepar'). Do not add class_filter unless the question is about a drug class, and never combine the two.
4. Flowsheet measurement names must be discovered, not guessed: call query_flowsheet with NO disp_name_filter first to see the exact names, then filter with a short substring taken from that list (e.g. 'CAM' rather than 'CAM-ICU', 'Temp' rather than 'Temperature').
5. Outputs are capped. If n_events_returned < n_events_total, n_rows_returned < n_rows_total, or a "_truncated" note appears, narrow the filter or time window (or raise max_rows) before drawing any conclusion from a partial list.
6. Prefer totals the tools already computed (per_label_minutes_active, n_events_total, n_rows_total) over adding long arrays by hand. If you must count or add over a long list, do it in blocks of 10 and keep a running total.
7. For what the patient was doing over a several-minute window, first compute the hour_index, then use get_atomic_actions for that hour (minute-level labels) together with the hourly report; use query_video_minute to check the specific minute (and the adjacent minute) when the question is about a single moment.
8. Investigate with at least two different tools or queries before answering whenever the tools can bear on the question, and cross-check them against each other.
9. Keep the reasoning short and end with the line: Answer: <LETTER>
\end{lstlisting}
\end{promptbox}
\begin{promptbox}[colback=gray!6]{Agent system prompt, text-only configuration}
\begin{lstlisting}[style=prompt]
You are answering a multiple-choice question about an ICU patient. You do NOT have access to the patient's video, activity data, or medical records, and you have no tools. Answer as best you can from general reasoning. You MUST select exactly one option as your single best guess, even if you are unsure or the data is incomplete. Never refuse and never say you cannot determine the answer.
\end{lstlisting}
\end{promptbox}
\begin{promptbox}[colback=gray!6]{User message for MCQ evaluation}
\begin{lstlisting}[style=prompt]
QUESTION: {question}

OPTIONS:
{options, one per line as 'A. ...'}

You must choose exactly one option even if uncertain -- never refuse.
End your reply with a line in exactly this form:
Answer: <LETTER>
\end{lstlisting}
\end{promptbox}
\begin{promptbox}[colback=gray!6]{Forced-choice message (sent when the 25-call budget is reached)}
\begin{lstlisting}[style=prompt]
You have used all available tool calls. Based on what you have found so far, commit to your single best-guess answer now. End with a line: Answer: <LETTER>
\end{lstlisting}
\end{promptbox}

\paragraph{Video tool.} \texttt{query\_video\_minute} sends the original one-minute clip of each camera view along with the same agent-written prompt, to the perception VLM (the agent's own model unless a row states otherwise) at 5~fps for Gemini models and 2~fps for open-weight models. A fixed preamble precedes the agent's prompt on every call, and both views' answers are returned for the agent to reconcile. The tool's full description, which the agent reads through the tool definition, is the expanded \texttt{query\_video\_minute} entry in the system prompt box above; the box below is the fixed preamble the tool prepends to the agent's question before sending the clip.

\begin{promptbox}[colback=green!4]{Preamble prepended to the agent's prompt on every video tool call}
\begin{lstlisting}[style=prompt]
You are an expert ICU video analyst. Patients here are often sedated, so subtle patient movement (a brief head turn, eye flutter, blink, cough, swallow, finger/limb twitch) is clinically relevant. Review the one-minute clip carefully and list ONLY movements you can actually see, each with an approximate timestamp (mm:ss). Do NOT infer or assume movements that are not clearly visible -- if the patient appears still, say so explicitly. Then answer the custom prompt below.
\end{lstlisting}
\end{promptbox}

\clearpage
\subsection{Temporal Hierarchy Construction Details}
\label{app:minute}
The perception VLM receives the 60 frames of a one-minute clip (1~fps) and a fixed prompt that requests a JSON object containing, for each of the 15 atomic actions, one of \textsc{yes}, \textsc{no}, or \textsc{unable to identify}, together with a brief free-text caption of the minute. The two camera views are merged by a logical OR over the 15 labels, with \textsc{unable to identify} treated as negative, giving one binary label vector per patient-minute and the two view-specific captions are kept. Storing each minute as a 15-bit vector enables efficient timeline tools and quick scanning of multi-day stays. So, we collapse \textsc{unable to identify} because it is relatively rare: Gemini 3.1 Flash Lite used it for 1 of 29{,}775 labels across five benchmark runs and in zero instances  of 198{,}810 labels in a uniform sample of 13{,}254 minute records from the cohort run. On the benchmark an abstained prediction is excluded from the metrics rather than scored as negative. Among other perception VLMs, only Gemini 2.5 Flash and 2.5 Pro abstain often (3\% of labels). Every other evaluated model abstains on under 0.5\%. Sampling settings for every model are listed in Appendix~\ref{app:sampling}.

\paragraph{Perception prompt.} The prompt below is sent with the 60 frames to every perception VLM, on the clip benchmark and on the cohort run (the clip identifier in the example is replaced by a placeholder). The user turn that follows it reads ``Please find the video frames below:'' and is followed by the frames.

\begin{promptbox}[colback=green!4]{Perception prompt (system turn)}
\begin{lstlisting}[style=prompt]
MAIN INSTRUCTIONS:
Your objective is to analyze video frames extracted from a 1-minute Intensive Care Unit (ICU) hospital patient surveillance video recorded by a fixed, corner-mounted camera.
This is STRICTLY a FINE-GRAINED video understanding task.
You will receive 60 images sampled every 1 second, totaling 1 minute.
Your ONLY task is to decide, for each predefined category below, whether the corresponding patient action or patient-clinician interaction is present in the video. Then write a short 1-2 sentence summary of this video (STRICTLY keep this short) summary.
You must ONLY output a structured JSON object with categorical labels and Short summary.
KEY VISUAL TASKS (TAXONOMY)
PATIENT MOTION -- ARMS/HANDS
  1) Grabbing medical equipment
     - The patient is actively grasping or holding medical devices
       (e.g., ventilator mask, tubing, IV lines, monitor cables).
  2) Using or grabbing non-medical objects
     - The patient is using or holding non-medical items
       (e.g., cup, towel, tissue, phone, food, remote).
  3) Gesticulating
     - The patient is making expressive arm or hand movements while communicating
       (e.g., waving, pointing, hand gestures while speaking).
  4) Adjusting pillows or bedsheets
     - The patient pulls, pushes, or rearranges pillows, blankets, or sheets.
  5) Other arm movements
     - Any noticeable arm/hand motion such as fidgeting, lifting/dropping arms,
       scratching, or non-specific movements.
     - This category can co-occur with other arm/hand categories above.

PATIENT MOTION -- TORSO
  6) Patient changes body position in bed
     - The patient changes posture in bed
       (e.g., lying flat to leaning on one side, sitting up to lying down).
  7) Other torso movement
     - Other significant torso movements
       (e.g., quick or repeated body movements, jerking, agitation-like movements).

PATIENT MOTION -- HEAD/NECK
  8) Patient is looking toward other people in the room
     - The patient's gaze is clearly directed at another person.
  9) Patient turns or moves head/neck
     - Visible head or neck movement
       (e.g., turning left/right, nodding, tilting up/down).

PATIENT MOTION -- LEGS
  10) Leg movements related to mobility
      - The patient is making leg motions associated with activities of daily living
        (e.g., standing up, sitting down, walking, transferring to or from the bed).
  11) In-bed leg movement
      - The patient is making leg movements while they remain in bed
        (e.g., leg fidgeting, repositioning, leg movements under the covers).

PATIENT-CLINICIAN INTERACTIONS
  12) Clinician touches patient or uses tool/device on patient
      - A clinician or other person makes physical contact with the patient
        OR applies a tool or device directly to the patient's body.
  13) Patient communication
      - The patient appears to be speaking, mouthing words, or nodding in a way that
        responds to another person.
  14) Object exchange
      - An item is being handed from a person to the patient or from the patient
        to another person (e.g., cup, towel, tool, brush, phone).
  15) Patient is being physically restrained or stabilized
      - Another person is clearly holding or stabilizing part of the patient's body
        in a way that limits or controls movement.

DECISION RULES
- Each category is INDEPENDENT. Multiple categories can be YES simultaneously
  for the same video. For example, a patient may be grabbing medical equipment
  (cat 1) while also performing other arm movements (cat 5) at the same time.
  Evaluate each category on its own merits.
- Base ALL judgments only on what is clearly visible in the frames. - Do NOT infer pain, emotions, intentions, diagnoses, or clinical reasoning. - If visibility is poor or you are not confident, use "UNABLE TO IDENTIFY" for that label. - Every label MUST be exactly one of:
    - "YES"
    - "NO"
    - "UNABLE TO IDENTIFY"
- Do NOT add extra fields or narrative text.
OUTPUT RULES
- Return a single JSON object with fixed keys for the video ID and all labels. - Do NOT include explanations, timestamps, or free-text descriptions. - Use the exact key names and allowed values shown below.
EXAMPLE OUTPUT FORMAT

  {
    "video_id": "<date>/<time>_<room>_lt.mp4",
    "labels": {
      "patient_grabbing_medical_equipment": "NO",
      "patient_using_non_medical_objects": "NO",
      "patient_gesticulating": "NO",
      "patient_adjusting_bedding": "NO",
      "patient_other_arm_movements": "NO",
      "patient_position_change": "NO",
      "patient_other_torso_movement": "NO",
      "patient_looking_at_others": "NO",
      "patient_head_turning": "NO",
      "patient_leg_ADL_movement": "NO",
      "patient_leg_in_bed_movement": "NO",
      "clinician_patient_contact": "YES",
      "patient_communication": "NO",
      "clinician_patient_object_exchange": "NO",
      "patient_restrained": "NO"
    },
    "summary": "A clinician is present and makes physical contact with the patient, while the patient remains mostly still with no notable movements."
  }

You must ALWAYS follow this structure exactly.
\end{lstlisting}
\end{promptbox}

\paragraph{Hourly report prompt.} The report model receives the system turn below, then a user turn that fixes the six-section format, the hour's 60 minute-level records from both camera views, and the medication administration events of that hour.

\begin{promptbox}[colback=blue!5]{Hourly report prompt (system turn)}
\begin{lstlisting}[style=prompt]
You are an ICU behavioral summarization model.
You will receive up to 60 minutes of Task-1 video caption outputs (one JSON entry per minute)
from two camera views (left and right) of an ICU patient room,
plus a list of medication administration events relevant to this hour.

Your job is to generate a structured behavioral report for the
observation period based purely on visible evidence from both camera views.
You must never infer sedation level, intent, pain, emotion, or clinical state.

When both camera views are available for a given minute, reconcile them:
if either view detects an action, treat it as observed. Note discrepancies
only when they are informative (e.g., one view blocked by equipment).

When only one camera view is available, note which view and proceed.

TIMING REQUIREMENT (IMPORTANT):
  - All behavioral findings must cite timing.
  - Use minute timestamps (e.g., min 14, min 27) AND/OR continuous ranges (e.g., min 10-15).
  - No event may be mentioned without a timing reference.

ALLOWED TIMING FORMATS:
  - "arm movement observed at min 12"
  - "clinician contact min 21-24"
  - "no movement detected min 0-18"

Your output must follow the 6-section template provided in the user prompt EXACTLY.
\end{lstlisting}
\end{promptbox}
\begin{promptbox}[colback=blue!5]{Hourly report prompt (user turn, before the hour's records)}
\begin{lstlisting}[style=prompt]
Generate the behavioral report for this observation hour.

REQUIRED FORMAT:

Observation window: [START_TIME] to [END_TIME] (up to 60 minutes)

1. Patient Movement Summary
    - Overall movement level (high / moderate / low / none)
    - Most common movement types
    - MUST cite minute timestamps or ranges for all movement observed
    - Note any discrepancies between left and right camera views

2. Patient-Clinician Interaction Summary
    - Whether interaction occurred
    - MUST cite timings for clinician contact, communication, restraint, exchanges

3. Significant Moments of Interest
    - Sudden spikes in movement, first movement after stillness, notable behavior
    - MUST include timing (single minute or range)
    - If prolonged stillness >= 10 min, cite the stillness window

4. Medication Context
    - List medications active or administered during this hour with timestamps
    - Include drug name, dose, route, and action (Given/Rate Change/New Bag/etc.)
    - Note any continuous infusions that are running

5. Activity-Medication Alignment (observational only)
    - Timing-based associations ONLY (e.g., "movement at min 22-28 while dexmedetomidine infusing")
    - No clinical cause/effect interpretation ever

6. Brief Summary (<= 5 sentences)
    - Compact overall description of the hour
    - Note uncertainty if many labels = "UNABLE TO IDENTIFY"
    - Note if camera coverage was incomplete

Output MUST contain all 6 sections and MUST cite times when describing events.
\end{lstlisting}
\end{promptbox}

\clearpage
\subsection{\mcqbenchmark{} Question Creation and Filtering}
\label{app:question-creation}

We generated MCQs through a three-stage process. In the first stage, questions were created manually by human annotators with access to the underlying patient data. The question types were informed by discussions with clinicians, nurses, and domain experts in long-form video understanding. Each question was designed to assess one of three capabilities: (1) understanding changes in patient status over extended periods of time ($>1$ hour), (2) understanding fine-grained patient activities over shorter time spans (1 minute--1 hour), or (3) retrieving and reasoning over electronic health record (EHR) information to answer clinical queries accurately.

In the second stage, the question text was revised to increase their difficulty, and plausible but incorrect distractor answers were created. These distractors were designed to reduce the effectiveness of superficial linguistic cues and discourage models from exploiting language-based biases instead of reasoning over the underlying patient data.

Finally, we manually verified the correctness of each question and corresponding distractors. Any question that blind LLMs with no patient data answered correctly with at least $60\%$ average accuracy was flagged for review, and flagged questions were dropped or reworked. Even after reworking, a number of questions remain guessable for the blind LLM baseline: with no patient data, 25 of the 144 questions (17\%) are answered correctly at least 80\% of the time by Gemini 3.1 Flash Lite. These questions account for part of the text-only baselines' accuracy above chance.

The final set of questions in \mcqbenchmark{} span temporal resolutions from concrete moments (``Around [time], what examination did the clinicians perform?'') to whole days (``During the first 24 hours of the stay, did the patient leave her bed or her room?''), and also test reasoning over the record (``How many times was the patient formally assessed for delirium during the stay?''). 

\paragraph{Evidence types.} For analysis every question is tagged by the evidence it requires. The 110 \emph{video} questions are split by the span of video that resolves them: \emph{moments} that a small number of clips can answer (under five minutes; 33 questions), \emph{windows} of 5 to 30 minutes (16), and \emph{long-form} spans of more than 30 minutes up to the whole stay (61).  The remaining questions are answerable from the record alone (\emph{EHR}; 26) or need both video and record to correctly answer (\emph{cross-modal}; 8). The benchmark probes many questions over two long trajectories and does not test generalization across patients. We note that accuracy differs between the two stays, so Appendix~\ref{app:ablation-models} reports each stay (denoted A and B) separately. 

\paragraph{Temporal anchoring.} Table~\ref{tab:time-anchor} splits the questions by the time reference they give: 60 name a clock time, 34 an hour, part of a day, or a day, and 50 give none. The agent localizes events through a variety of mechanisms largely depending on the information present in the question: whether a time is provided in the question, whether it mentions a specific event present in the EHR, whether the question is able to be localized by an action query over the timeline, or by simply reading hourly reports over selected windows. The most specific temporal references generally belong to video-moment questions, where the difficulty generally lies in perception and not localization.

\paragraph{Agent prompt development.} We refined the agent prompt using earlier draft versions of benchmark questions. None of the final 144 questions, in their evaluated form, was used for prompt development. All reported evaluations used a single fixed agent prompt.

\begin{table}[h]
  \centering
  \footnotesize
  \caption{\textbf{\mcqbenchmark{} question breakdown by time reference.} Accuracy (\%) for Gemini 3.1 Flash Lite using \method{}, from an independent run of the Table~\ref{tab:mcq-performance} configuration (achieves 61.1\% overall).}
  \vspace{2mm}
  \label{tab:time-anchor}
  \begin{tabular}{@{} l r r @{}}
    \toprule
    Time reference in the question & Questions & Accuracy \\
    \midrule
    Clock time, span under 5 minutes & 33 & 30.9 \\
    Clock time, span 5--30 minutes & 15 & 57.3 \\
    Clock time, span over 30 minutes & 12 & 50.0 \\
    Hour & 13 & 50.8 \\
    Part of day (e.g., morning, afternoon) & 16 & 66.2 \\
    Day (e.g., November 23) & 5 & 56.0 \\
    No explicit time, anchored on clinical record & 22 & 96.4 \\
    No explicit time, anchored on video data & 28 & 78.6 \\
    \bottomrule
  \end{tabular}
  \vspace{-3mm}
\end{table}

\clearpage
\subsection{Patient Cohort}
\label{app:cohort}

\begin{figure}[!ht]
    \begin{minipage}[t]{0.42\linewidth}
        \vspace{1mm}
        \includegraphics[width=1.4\linewidth]{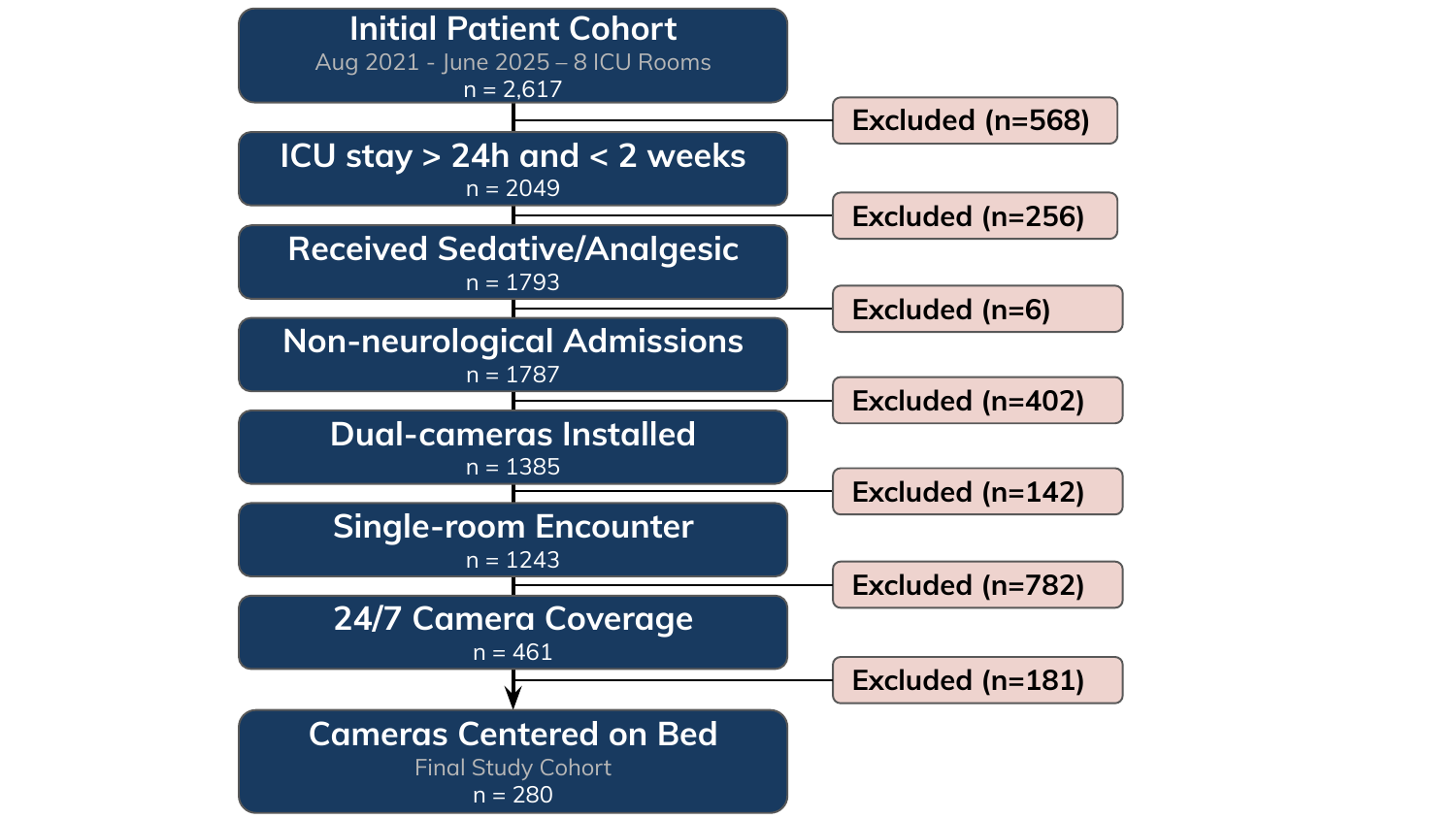}
    \end{minipage}\hfill
    \begin{minipage}[t]{0.55\linewidth}
        \centering
        \vspace{6pt}
        \scriptsize
        \begin{tabular}{lc}
            \toprule
            \textbf{Characteristic} & \textbf{Overall ($N = 280$)} \\
            \midrule
            Age, years, median [IQR]                         & 66.0 [58.1, 74.7] \\
            Male, $n$ (\%)                                   & 183 (65.4) \\
            BMI, kg/m$^2$, median [IQR]                      & 25.8 [23.2, 29.9] \\
            \addlinespace
            \multicolumn{2}{l}{\textit{Admission type, $n$ (\%)}} \\
            \quad Elective                                   & 140 (50.0) \\
            \quad Emergency                                  & 103 (36.8) \\
            \quad Urgent                                     & 30 (10.7) \\
            \quad Trauma Center                              & 7 (2.5) \\
            \addlinespace
            \addlinespace
            ICU LOS, days, median [IQR]                      & 3.3 [2.0, 5.4] \\
            Invasive mechanical ventilation, $n$ (\%)        & 157 (56.1) \\            %
            In-hospital mortality, $n$ (\%)                  & 13 (4.6) \\
            CAM-ICU positive during stay, $n$ (\%)          & 53 (18.9) \\
            \bottomrule
        \end{tabular}
    \end{minipage}

    \caption{\textbf{Patient cohort selection and characteristics.} \textbf{Left}, CONSORT-style flow diagram of patient cohort selection. Sequential exclusion criteria were applied to identify patients with sufficient ICU stay duration, sedative exposure, non-neurological admission, and adequate bilateral video coverage. \textbf{Right}, Characteristics of the final study cohort ($N = 280$).}
            \label{fig:cohort}

\end{figure}

We visualize our clinical cohort and characteristics of our final 280 patient study cohort in Figure \ref{fig:cohort}. 

\subsubsection{Patient Cohort Selection}
From the initial pool of 2,617 patients assessed for eligibility between August 2021 and June 2025, we applied sequential exclusion criteria to identify a cohort suitable for evaluating continuous video-based behavioral monitoring. Patients were required to have ICU stays exceeding 24 hours but less than two weeks (excluding 568), have received at least one sedative or analgesic medication (excluding 256), and have non-neurological admitting diagnoses (excluding 6). Technical requirements included admission to a room with two cameras  installed (excluding 402), single-room encounters with no intra-ICU transfers (excluding 142), continuous bilateral video feed with 100\% coverage over the entire stay (excluding 782), and confirmation that both cameras were centered on the patient bed area (excluding 181). See Figure~\ref{fig:cohort} for a visualization.

\subsubsection{Electronic Health Records}

The harness reads four kinds of structured records, each through a specific component (tool signatures in Appendix~\ref{app:tools}). (1) \textit{Demographics and encounter metadata}: age, sex, race and ethnicity, height and weight, and per-encounter fields (admission type and source, hospital service, department and room, level of care, ICU days, discharge disposition), returned to the agent by the patient-metadata tool together with the ICU admission time and the video observation window. (2) \textit{Admission, discharge, and transfer (ADT) events}: used to define ICU stays as contiguous periods at the ``Critical Care'' level of care, room assignments, and the observation window; the agent sees them only through the windows and room reported by the metadata tool. (3) \textit{Medication administration records}: timestamped administrations with drug name, dose, route, action (Given, Rate Change, New Bag, Stopped, and similar), and pharmacologic and therapeutic class, exposed by the medication tool and supplied to the report model as the medication context of each hour. (4) \textit{Nursing flowsheet measurements}: timestamped rows with a measurement name, value, and unit, covering vital signs (heart rate, blood pressure, oxygen saturation, respiratory rate), sedation and agitation scores (Richmond Agitation-Sedation Scale, anxiety, pain scales), Confusion Assessment Method for the ICU (CAM-ICU) delirium screens, and nursing assessments (activity, activities of daily living, communication, cardiac rhythm), exposed by the flowsheet tool, whose discovery mode lists the measurement names present for a patient. The EHR questions of \mcqbenchmark are answerable from (3) and (4); the cohort analysis of Section~\ref{sec:results-cohort} takes its delirium labels from the CAM-ICU screens in (4) and the cohort characteristics of Table~\ref{tab:delirium-cohort} from (1), (2), and (4).

\clearpage
\subsection{Token Cost of Raw Video}
\label{app:tokens}
Table~\ref{tab:tokens} quantifies the coverage--detail trade-off described briefly in Section~\ref{sec:problem}. For the one-minute clip the Gemini video tool sends the original 1920$\times$1080 recording with 5~fps sampling. This results in 19{,}823 tokens per view, identical for both Gemini 3.1 Flash Lite and Gemini 3.5 Flash. We use 950 frames at 512$\times$288 sampled uniformly across the stay for the long-context baseline, which uses 1{,}100 tokens per frame. For Qwen3-VL 8B, the same inputs were counted (patch size 16, $2\times2$ spatial merge, temporal patch size 2, frames used at native size): 144 tokens per individual frame and 8{,}640 tokens per one-minute clip. Under either tokenizer a single day of one camera exceeds the model's context by more than an order of magnitude. The 950-frame prompt still exists at the edge of a million-token window and, spread over a multi-day stay, samples one frame every several minutes. Thus, most events shorter than the sampling interval are likely to be missed.  Table~\ref{tab:mcq-performance} shows the consequence, with every long-context baseline without EHR only able to land between 28.6\% and 33.0\%. By constructing the hierarchy as the stay is recorded and not using the entire hierarchy in context, each question requires a significantly smaller number of tokens. 

\begin{table}[h]
  \centering
  \small
  \caption{\textbf{Token cost of raw video surpasses context lengths.} Context windows: 1M tokens for the Gemini models, 256K for Qwen3-VL 8B. Gemini uses the original 1920$\times$1080 clip at 5~fps. Qwen3-VL uses 512$\times$288 resolution at 2~fps. }
  \label{tab:tokens}
  \vspace{1mm}
  \begin{tabular}{@{}lrr@{}}
    \toprule
    Input & Gemini 3 Family (tokens) & Qwen3-VL 8B (tokens)\\
    \midrule
    One minute, one camera view  & 19{,}823 & 8{,}640 \\
    One minute, both views & 39{,}646 & 17{,}280 \\
    One hour, both views & 2.38M & 1.04M \\
    One day, both views & 57.1M & 24.9M \\
    Median stay (3.3 days) & 188M & 82M \\
    Longest stay in cohort (14 days) & 799M & 349M \\
    \mcqbenchmark{} span (264 h) & 628M & 274M \\
    \midrule
    One 512$\times$288 frame & 1{,}100 & 144 \\
    950 stills over a stay (long-context baseline) & 1.05M & 0.14M \\
\bottomrule
  \end{tabular}
\end{table}

\clearpage
\subsection{Baseline Implementations}
\label{app:baselines}
All baselines answer the same number of questions (5 gold answer placements per question) as the harness and are scored identically. Two of the 144 questions have four options rather than five, so each configuration is scored on 718 trials. The confidence interval in Section~\ref{sec:results-qa} averages each question's placements, resamples the 144 questions with replacement (10{,}000 percentile-bootstrap resamples, the same questions for both systems); the McNemar tests in the appendix instead treat placements as independent trials and are therefore approximate. \paragraph{Text-only.} The text-only agent runs through the same agent loop with no tools and the short text-only system prompt of Appendix~\ref{app:tools}, which states that it has no access to the patient's data. Its accuracy is intended to measure what language priors recover.

\paragraph{Socratic Models.} Following the HourVideo protocol, a VLM captions every minute of the stay with the published HourVideo caption prompt, and a language model of the same family answers from the captions with the published HourVideo question prompt. This is a separate captioning pass from the harness's minute records: it uses a different prompt and produces free-text captions rather than action labels. Gemini 3.1 Flash Lite captions each clip at its default 1~fps and Qwen3-VL 8B is given 30 frames per clip. When the captions exceed the answering model's context they are subsampled at a fixed linear stride. For Gemini at 286 tokens per caption inside a million-token window, we sample every other minute for stay A (2{,}836 of 5{,}671 minutes) and every third for stay B (3{,}385 of 10{,}153). For the open-weight answerers of Table~\ref{tab:socratic-gemcap}, tokens per caption range from 301 to 320, resulting in captions for 183 to 391 minutes per stay depending on the model's available context. Gemini baselines run at API defaults; the open-weight baselines use greedy decoding in bfloat16 (Appendix~\ref{app:sampling}).

\paragraph{Socratic baselines with clinical records and both cameras.}
\label{app:socratic-ehr}
The Socratic methods described in the paragraph above use one camera and no clinical records, whereas \method uses both cameras and can query the record. Thus, we add two Socratic variants with the same backbone (Gemini 3.1 Flash Lite as captioner and answerer) that close those gaps within a single million-token prompt (Table~\ref{tab:socratic-ehr}). Both prepend an \emph{EHR block} to the captions: the patient metadata, every medication administration event, and every flowsheet variable of the stay. This block costs 131{,}000 tokens for stay A (670 flowsheet variables) and 420{,}000 for stay B (835 variables). The caption stride is re-derived so that block, captions, and prompts fit the model's context window. The first variant uses only the left-camera captions: every second minute survives for stay A, as before, and every sixth for stay B, resulting in prompts of 998{,}000 and 939{,}000 tokens. The second adds a right-camera caption for every minute, produced by the same prompt, model, and frame rate (15{,}824 clips), and keeps both captions of each surviving minute. This results in every fourth minute caption used for stay A and every eleventh for stay B and gives prompts of 689{,}000 and 706{,}000 tokens. Both prompts are several times the harness's per-question budget. This approach improves the performance on EHR questions from 16.2 to 76.2 and the average from 33.7 to 45.0 (McNemar over the 718 paired trials, $p<0.001$), with video performance unchanged. Adding the second camera does not help further at a fixed budget (42.7) due to the larger temporal stride. Performance falls on long-form video questions (43.3 to 35.9) and moment and cross-modal questions move within noise. The harness with the same model reaches 61.0.

\paragraph{Long-context VLM.} The model receives the question together with 475 or 950 frames under a fixed frames-only prompt. Frames are spaced uniformly over the minute clips of the stay, taking the midpoint of each selected clip, and are extracted from the left camera (the right camera is used only for minutes where the left clip is missing) and scaled so that the longer side is 512 pixels, giving 512$\times$288. Frames are given in chronological order without timestamps. On Gemini 3.1 Flash Lite the 950-frame prompt uses 1{,}045{,}000 tokens (1{,}100 per frame), hitting the limit of its million-token window; 475 is half of that, and on Qwen3-VL 8B the same frames cost 144 tokens each. Over the two benchmark stays, 950 frames correspond to one frame every 6.0 and 10.7 minutes, so events shorter than that interval are likely to be missed by this baseline.  

\paragraph{Long-context baseline with the clinical record.} The same EHR block described above is added to the long-context baseline and we halve the frame count, resulting in 475 frames at 512$\times$288 resolution from the left camera. The frames follow the EHR block in a prompt spanning 654{,}000 tokens for stay A and 943{,}000 for stay B. This is the strongest baseline and scores 49.3 (Table~\ref{tab:socratic-ehr}). Adding the EHR to the 475 frame baseline lifts long-form video performance from 36.6 to 48.9 as well as EHR performance from 17.7 to 83.8.

\begin{table}[h]
  \centering
  \small
  \caption{\textbf{Baselines with clinical records using Gemini 3.1 Flash Lite.} Accuracy (\%) averaged over gold placements. ``Coverage'' is the number of captioned minutes or frames that survive subsampling given the fixed 1M context for stay A / stay B; ``Tokens'' is the prompt length per trial. \method is shown on the last row for comparison.}
  \label{tab:socratic-ehr}
  \vspace{1mm}
  \resizebox{\linewidth}{!}{%
  \begin{tabular}{@{}lrrcccccc@{}}
    \toprule
    Inputs & Coverage & Tokens & $<$5 min & 5--30 min & $>$30 min & EHR & Cross & All \\
    \midrule
    Left camera captions & 2.8k / 3.4k & 1.0M & 34.5 & 32.5 & 39.2 & 16.2 & 47.5 & 33.7 \\
    Left camera captions + EHR  & 2.8k / 1.7k & 1.0M / 0.9M & 30.3 & 32.5 & 43.3 & 76.2 & 42.5 & 45.0 \\
    Both cameras' captions + EHR  & 1.4k / 0.9k & 0.7M / 0.7M & 32.7 & 31.2 & 35.9 & 76.2 & 50.0 & 42.7 \\
    \addlinespace[2pt]
    950 frames & 950 / 950 & 1.1M & 35.8 & 32.5 & 38.0 & 14.6 & 25.0 & 31.9 \\
    475 frames & 475 / 475 & 0.5M & 26.7 & 35.0 & 36.6 & 17.7 & 22.5 & 29.9 \\
    475 frames + EHR  & 475 / 475 & 0.7M / 1.0M & 34.5 & 30.0 & 48.9 & 83.8 & 40.0 & 49.3 \\
    \midrule
    \method{} harness, all tools (Table~\ref{tab:mcq-performance} run) & -- & -- & 36.4 & 52.5 & 63.7 & 96.2 & 45.0 & 61.0 \\
    \bottomrule
  \end{tabular}}
\end{table}

\paragraph{Open-weight question answering for Socratic baselines.}  Table~\ref{tab:socratic-gemcap} runs the Socratic protocol with the Gemini 3.1 Flash Lite captions for video perception and an open-weight model as the answerer. For each row, we show the performance of \method using the same model as the agent LLM.  Every open-weight answerer lands within about ten points of chance when it can only read the captions, and gains 33 to 39 points when using the harness. 

\begin{table}[h]
  \centering
  \small
  \caption{\textbf{Socratic baselines with Gemini 3.1 Flash Lite captions, by open-weight answering model.} Accuracy (\%) averaged over gold placements; ``minutes'' is the number of captioned minutes that survive subsampling for stay A / stay B; the last column is the same model as the \method agent with Gemini 3.1 Flash Lite serving the video tool. Served context length shown in parentheses.}
  \label{tab:socratic-gemcap}
  \vspace{1mm}
  \begin{tabular}{@{}lrccccccc@{}}
    \toprule
    Answering model & Minutes & $<$5 min & 5--30 min & $>$30 min & EHR & Cross & All & \method \\
    \midrule
    Gemini 3.1 FL (1M) & 2.8k / 3.4k & 34.5 & 32.5 & 39.2 & 16.2 & 47.5 & 33.7 & 61.0 \\
    gpt-oss-20b (131K) & 379 / 391 & 33.3 & 26.3 & 22.7 & 14.6 & 20.0 & 23.9 & 57.4 \\
    Gemma-4 31B (131K) & 379 / 377 & 33.9 & 25.0 & 31.4 & 23.1 & 22.5 & 29.3 & 67.9 \\
    Qwen3.5 9B (65K) & 183 / 185 & 30.3 & 28.8 & 30.8 & 18.5 & 10.0 & 27.1 & 61.6 \\
    \bottomrule
  \end{tabular}
\end{table}

\clearpage
\subsection{Model Instantiation and Computational Cost}
\label{app:models}

All three roles are frozen, prompted models; their sampling settings are listed at the end of this section. For temporal abstraction we use Gemini 3.1 Flash Lite as both the perception and the report model. Although it is not the most accurate VLM we evaluate, it offered the best accuracy--cost trade-off for running over roughly 60{,}000 camera-hours. Hourly reports were generated, with the same model, only for the two \mcqbenchmark{} stays. To measure how perception quality varies across models we evaluate nine VLMs on the perception benchmark under the same prompt: the open-weight Qwen3-VL 8B~\citep{bai2025qwen3}, Qwen3.5 4B and 9B~\citep{team2026qwen3}, and Gemma-4 E4B~\citep{team2026gemma4}, and five Gemini models spanning three generations~\citep{geminiteam2023gemini,comanici2025gemini}; our data use agreement permits processing protected health information among proprietary models only within the Gemini family. For the agent model we report Gemini 3.5 Flash as the best performing model, and generally use Gemini 3.1 Flash Lite as the reference configuration for most baseline comparisons.

\paragraph{Computational cost.} Almost all visual computation happens in the background: constructing the temporal hierarchy costs two VLM calls per minute of stay and one LLM report call per hour. This cost is independent of how many questions are later asked. Answering a question is cheap by comparison. For Gemini 3.5 Flash, we find that the agent uses an average of 160{,}000 tokens per question over 9.8 turns. On average, each question uses 3.3 video calls adding approximately 132{,}000 tokens, resulting in roughly 292{,}000 per question. The distribution is heavy-tailed because a few trials reach the 25-call cap with long transcripts.

\paragraph{Sampling and serving settings.}
\label{app:sampling}
Gemini models are called with the API defaults for temperature, top-$p$, top-$k$, output length, and thinking in every role (perception, reports, agent, and baselines). Open-weight models are served with vLLM with thinking disabled. For the atomic action benchmark we use temperature 0.6, top-$p$ 0.95, and top-$k$ 20. For the long-context and Socratic baselines we use greedy decoding. Every agent runs under the same loop: the agent prompt, at most 25 tool calls, a forced answer when the budget is exhausted. The video tool sends the original one-minute clip with  5~fps sampling for Gemini models and 512$\times$288 at 2~fps for open-weight models. The cohort perception run and the hourly reports use Gemini 3.1 Flash Lite at API defaults with the prompts of Appendix~\ref{app:minute}.

\clearpage
\subsection{Perception Benchmark Details}
\label{app:cost}

\begin{figure}[h]
  \centering
  \includegraphics[width=0.98\linewidth]{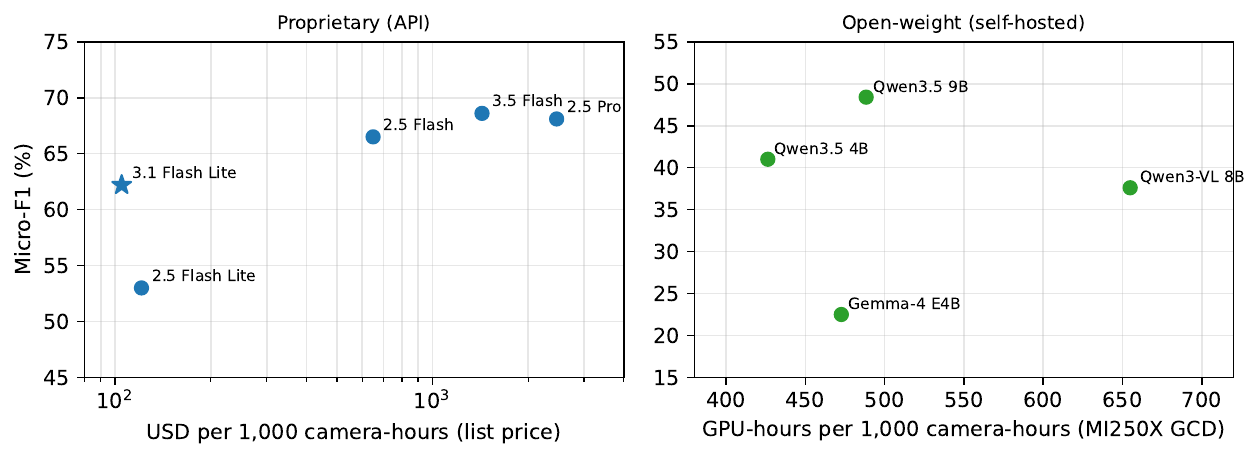}
  \caption{\textbf{Perception accuracy versus cost.} Left: Gemini models, list-price cost per 1,000 camera-hours of one-minute clips (log scale); the star marks the model used for the cohort run. Right: open-weight models, device-hours per 1,000 camera-hours on one MI250X graphics compute die.}
  \label{fig:cost}
\end{figure}

\paragraph{Accuracy-cost tradeoff.}
Figure~\ref{fig:cost} plots micro-F1 on the 397-clip benchmark against the cost of running each perception VLM over video. For the Gemini models, cost is computed from all recorded prompt and response tokens using the prices at the time of submission (September 2026). For the open-weight models, cost is measured wall-clock per clip on one AMD MI250X graphics compute die. Gemini 3.1 Flash Lite was selected to serve as the perception VLM for many of our experiments due to its strong and cheap performance at the Pareto frontier. 

\paragraph{Prevalence.}
In our 397-clip benchmark, category prevalence is imbalanced: the two most common actions (other arm movements, head turning) are present in $\sim$70\% of the 397 clips, while the rarest appear in fewer than 5\%.  We note that the relative prevalence of atomic actions in the actual ICU video data is significantly lower for all categories due to our selection of clips with differing levels of patient activity, acuity, and time of day throughout the ICU stay.

\paragraph{Annotation reliability.}
\label{app:reliability}
 In order to test annotation reliability, one author independently re-annotated 20 randomly sampled benchmark clips annotated by others, blinded to the original labels, model predictions, and clip metadata, using the finalized rubric. Agreement with the final verified benchmark labels, pooled across all 15 categories, was 92.7\%, with $\kappa = 0.81$ (95\% CI [0.74, 0.87], clip-level bootstrap; $N = 300$ paired decisions).

\clearpage
\subsection{Hourly Report Faithfulness}
\label{app:faithfulness}
Because the hourly reports are generated by a language model and then consumed as evidence by another, we audited ten reports, five per benchmark stay. Every claim that cites a minute, an action, a medication, or a temporal or causal relationship was extracted and checked against the minute-level records and the medication administration record. From our sample, we found 153 total claims (Table~\ref{tab:faithfulness}). Approximately 84\% of these claims were clearly supported by the minute records and EHR. Most of the remainder (8\%) were timing claims whose cited minutes only partly matched the minute records; six (4\%) were statements that no medication was given during an hour in which the record shows one; the rest (4\%) were overall movement ratings or absence claims that were contradicted by the minute records (e.g., describing an activity period as ``low activity'' despite many atomic action detections). Support was highest in the narrative summary and significant-moments sections and lowest in the movement and activity--medication sections. Every claim that named a medication matched the record on drug, dose, route, and minute. The dominant error was \emph{false absence}: three reports stated that no medication was administered during an hour in which the record shows one, and each such statement propagated into the activity--medication section. We found no invented causal language in the sample. We note the limitations of our approach: a claim-level audit cannot detect omission. A larger audit covering per-hour completeness across a larger number of patients is necessary before deployment.

\begin{table}[h]
  \centering
  \scriptsize
  \caption{\textbf{Claim-level audit of ten hourly reports.} Counts are verifiable claims. Partial timing match: under 80\% of cited minutes show the action. Carry over: a false medication absence repeated in the alignment section.}
  \label{tab:faithfulness}
  \setlength{\tabcolsep}{3pt}
  \begin{tabularx}{\linewidth}{@{} >{\raggedright\arraybackslash}X r r r r r r r r @{}}
    \toprule
    & \multicolumn{6}{c}{Report section} & \\
    \cmidrule(lr){2-7}
    & Movement & Interaction & Moments & Medication & Alignment & Summary & All & \% \\
    \midrule
    Verifiable claims & 41 & 36 & 28 & 16 & 12 & 20 & 153 & \\
    Supported & 31 & 32 & 25 & 13 & 9 & 19 & 129 & 84.3 \\
    \addlinespace[3pt]
    Errors & 10 & 4 & 3 & 3 & 3 & 1 & 24 & 15.7 \\
    \quad Only partial timing match & 7 & 1 & 3 & 0 & 0 & 1 & 12 & 7.8 \\
    \quad Movement summary misleading & 3 & 0 & 0 & 0 & 0 & 0 & 3 & 2.0 \\
    \quad Incorrect action absence & 0 & 3 & 0 & 0 & 0 & 0 & 3 & 2.0 \\
    \quad Incorrect medication absence & 0 & 0 & 0 & 3 & 0 & 0 & 3 & 2.0 \\
    \quad Carry over of incorrect information & 0 & 0 & 0 & 0 & 3 & 0 & 3 & 2.0 \\
    \midrule
    Support rate (\%) & 76 & 89 & 89 & 81 & 75 & 95 & 84 & \\
    \bottomrule
  \end{tabularx}
\end{table}

\clearpage
\subsection{How does \method{} Perform Across Different Agent Models?}
\label{app:backbones}
Table~\ref{tab:backbones} runs the full \method{} harness across 12 agent configurations. We find that Qwen3.5 9B's accuracy increases from 57.9 to 61.6 when using Gemini 3.1 Flash Lite for video perception.  Gemma-4 31B is the strongest open-weight agent at 67.9, saturating the EHR questions and reaching 73.7 for long-form video questions. Gemini 3.5 Flash with Gemini 3.1 Flash Lite serving its video tool scores 70.7 against its original 76.9 (paired McNemar on 718 shared trials, 103 versus 59 discordant, $\chi^2=11.4$, $p<0.001$), with the loss confined to video moments (54.5 to 35.8) and windows (75.0 to 57.5) while long-form and EHR questions are unchanged. This weaker perceiver also makes the agent re-query video more, hitting the 25-call budget on 10.7\% of trials against none with its own perception.

\begin{table}[h]
  \centering
  \small
  \caption{\textbf{\method applied across twelve agent configurations.} Full \method with the agent model as the only variable; the background records and reports are fixed. Tool calls are means per question. $^\dagger$Video tool served by Gemini 3.1 Flash Lite rather than the agent's own model.}
  \label{tab:backbones}
  \vspace{1mm}
  \begin{tabular}{@{}lccccccc@{}}
    \toprule
    Agent model & All & $<$5 min & 5--30 min & $>$30 min & EHR & Cross & Avg. Calls \\
    \midrule
    Gemini 2.5 Flash Lite   & 49.2 & 32.7 & 35.0 & 53.9 & 69.2 & 45.0 & 3.7 \\
    Gemini 2.5 Flash        & 60.3 & 33.9 & 56.2 & 64.0 & 90.8 & 50.0 & 4.7 \\
    Gemini 3 Flash          & 70.8 & 44.8 & 66.3 & 77.0 & 91.5 & 72.5 & 12.3 \\
    Gemini 3.1 Flash Lite   & 61.0 & 36.4 & 52.5 & 63.7 & 96.2 & 45.0 & 8.9 \\    Gemini 3.5 Flash Lite   & 61.1 & 40.0 & 52.5 & 59.8 & 99.2 & 52.5 & 6.8 \\
    Gemini 3.5 Flash        & \textbf{76.9} & \textbf{54.5} & \textbf{75.0} & \textbf{80.7} & 99.2 & 72.5 & 9.0 \\
    Qwen3.5 9B              & 57.9 & 26.7 & 43.8 & 63.3 & 95.4 & 52.5 & 10.4 \\
    Qwen3-VL 8B             & 42.1 & 19.4 & 32.5 & 45.7 & 68.5 & 42.5 & 4.7 \\
    gpt-oss-20b$^\dagger$   & 57.4 & 32.1 & 41.2 & 62.5 & 93.1 & 40.0 & 3.7 \\
    Qwen3.5 9B$^\dagger$    & 61.6 & 35.2 & 45.0 & 67.3 & 95.4 & 50.0 & 9.0 \\
    Gemma-4 31B$^\dagger$   & 67.9 & 40.0 & 61.2 & 73.7 & \textbf{100.0} & 47.5 & 5.4 \\
    Gemini 3.5 Flash$^\dagger$ & 70.7 & 35.8 & 57.5 & 80.3 & 98.5 & \textbf{77.5} & 9.5 \\
    \bottomrule
  \end{tabular}
\end{table}

\clearpage
\subsection{Per-Model Cumulative Ablation}
\label{app:ablation-models}
Table~\ref{tab:mcq-ablation-models} gives the cumulative ablation of Table~\ref{tab:mcq-ablation} across many agent models separately. 

\begin{table}[h]
  \centering
  \footnotesize
  \caption{\textbf{Per-model cumulative ablation, by question type and by stay.} Cumulative ablation: each row adds one tool family to the row above (Table~\ref{tab:tools}). Paired McNemar tests on identical trials for Gemini 3.5 Flash show that every step is significant ($p<0.001$) except adding reports ($+3.5$, $\chi^2=3.6$, $p=0.06$).}
  \label{tab:mcq-ablation-models}
  \vspace{1mm}
  \begin{tabular}{@{}l ccc c c cc c@{}}
    \toprule
    & \multicolumn{3}{c}{\cellcolor{green!18}\textbf{Video}} & \cellcolor{orange!22}\textbf{EHR} & \cellcolor{blue!14}\textbf{Cross-modal} & \multicolumn{2}{c}{\cellcolor{gray!15}\textbf{Stay}} & \textbf{All} \\
    Agent tools & $<$5 min & 5--30 min & $>$30 min & Record only & Video + EHR & A & B & \\
    \midrule
    \rowcolor{gray!12} \multicolumn{9}{@{}l}{\textit{Gemini 3.5 Flash}} \\
    Text-only & 37.0 & 32.5 & 51.4 & 21.5 & 27.5 & 35.9 & 41.6 & 39.3 \\
    + timeline & 37.6 & 57.5 & 67.5 & 42.3 & 20.0 & 62.0 & 45.6 & 52.4 \\
    + reports & 40.6 & 72.5 & 70.5 & 38.5 & 30.0 & 70.2 & 45.9 & 55.8 \\
    + EHR & 39.4 & 76.2 & 71.1 & 100.0 & 75.0 & 84.7 & 59.5 & 69.9 \\
    + video (\method) & \textbf{54.5} & 75.0 & \textbf{80.7} & 99.2 & 72.5 & \textbf{88.5} & \textbf{68.9} & \textbf{76.9} \\
    \midrule
    \rowcolor{gray!12} \multicolumn{9}{@{}l}{\textit{Gemini 3.1 Flash Lite}} \\
    Text-only & 24.2 & 18.8 & 40.2 & 26.9 & 25.0 & 26.8 & 33.8 & 30.9 \\
    + timeline & 23.0 & 32.5 & 55.3 & 26.2 & 22.5 & 50.5 & 29.8 & 38.3 \\
    + reports & 24.8 & 48.8 & 62.9 & 31.5 & 37.5 & 57.9 & 36.9 & 45.5 \\
    + EHR & 23.6 & 36.2 & 58.6 & 96.2 & 47.5 & 71.2 & 42.5 & 54.3 \\
    + video (\method) & 36.4 & 52.5 & 63.7 & 96.2 & 45.0 & 75.8 & 50.8 & 61.0 \\
    \midrule
    \rowcolor{gray!12} \multicolumn{9}{@{}l}{\textit{Qwen3.5 9B}} \\
    Text-only & 31.5 & 27.5 & 43.9 & 20.0 & 40.0 & 35.3 & 34.4 & 34.7 \\
    + timeline & 21.8 & 40.0 & 59.3 & 22.3 & 32.5 & 52.1 & 32.2 & 40.4 \\
    + reports & 26.1 & 45.0 & 58.0 & 23.8 & 27.5 & 51.9 & 34.0 & 41.4 \\
    + EHR & 32.1 & 42.5 & 62.5 & 95.4 & 62.5 & 74.8 & 48.4 & 59.2 \\
    + video (\method) & 26.7 & 43.8 & 63.3 & 95.4 & 52.5 & 74.6 & 46.4 & 57.9 \\
    \midrule
    \rowcolor{gray!12} \multicolumn{9}{@{}l}{\textit{Qwen3-VL 8B}} \\
    Text-only & 29.7 & 28.7 & 42.6 & 25.4 & 32.5 & 34.9 & 34.1 & 34.4 \\
    + timeline & 21.2 & 27.5 & 39.9 & 22.3 & 35.0 & 39.3 & 24.9 & 30.8 \\
    + reports & 19.4 & 48.8 & 46.7 & 18.5 & 22.5 & 39.5 & 30.6 & 34.2 \\
    + EHR & 17.0 & 33.8 & 43.5 & 70.0 & 25.0 & 53.8 & 30.6 & 40.1 \\
    + video (\method) & 19.4 & 32.5 & 45.7 & 68.5 & 42.5 & 55.7 & 32.7 & 42.1 \\
    \midrule
    \rowcolor{gray!12} \multicolumn{9}{@{}l}{\textit{gpt-oss-20b (video tool served by Gemini 3.1 Flash Lite)}} \\
    Text-only & 22.4 & 17.5 & 36.5 & 22.3 & 42.5 & 27.5 & 29.9 & 28.9 \\
    + timeline & 22.4 & 36.2 & 56.5 & 16.2 & 35.0 & 47.4 & 31.4 & 38.0 \\
    + reports & 25.5 & 38.8 & 62.1 & 27.7 & 30.0 & 55.8 & 34.3 & 43.1 \\
    + EHR & 26.7 & 40.0 & 59.3 & 90.8 & 52.5 & 74.8 & 41.2 & 55.0 \\
    + video (\method) & 32.1 & 41.2 & 62.5 & 93.1 & 40.0 & 73.4 & 46.4 & 57.4 \\
    \bottomrule
  \end{tabular}
\end{table}

\clearpage
\subsection{Single Tool Family Ablations}
\label{app:toolsalone}
Table~\ref{tab:toolsalone} gives the agent a single tool family on top of the scaffold tools, following the same protocol as Table~\ref{tab:mcq-ablation}.

\begin{table}[h]
  \centering
  \small
  \caption{\textbf{Single tool families on \mcqbenchmark.} Each row gives the agent one tool family only. Accuracy (\%) averaged over gold placements; category columns are question means.}
  \label{tab:toolsalone}
  \vspace{1mm}
  \begin{tabular}{@{}lcccccc@{}}
    \toprule
    Agent tools & All & $<$5 min & 5--30 min & $>$30 min & EHR & Cross \\
    \midrule
    \multicolumn{7}{@{}l}{\textit{Gemini 3.5 Flash}} \\
    Reports only & 51.0 & 44.2 & 72.5 & 55.4 & 36.9 & 47.5 \\
    EHR only     & 59.6 & 49.1 & 47.5 & 49.5 & 100.0 & 72.5 \\
    Video only   & 46.4 & 54.5 & 76.2 & 43.9 & 30.0 & 27.5 \\
    All tools    & 76.9 & 54.5 & 75.0 & 80.7 & 99.2 & 72.5 \\
    \midrule
    \multicolumn{7}{@{}l}{\textit{Gemini 3.1 Flash Lite}} \\
    Reports only & 37.5 & 27.9 & 38.8 & 44.5 & 31.5 & 40.0 \\
    EHR only     & 45.3 & 26.1 & 20.0 & 40.7 & 95.4 & 47.5 \\
    Video only   & 34.5 & 38.2 & 45.0 & 36.8 & 22.3 & 20.0 \\
    All tools    & 61.0 & 36.4 & 52.5 & 63.7 & 96.2 & 45.0 \\
    \midrule
    \multicolumn{7}{@{}l}{\textit{Qwen3.5 9B}} \\
    Reports only & 37.7 & 29.1 & 46.2 & 42.2 & 31.5 & 42.5 \\
    EHR only     & 43.0 & 23.6 & 27.5 & 33.5 & 93.8 & 60.0 \\
    Video only   & 30.2 & 26.7 & 31.2 & 34.3 & 23.1 & 35.0 \\
    All tools    & 57.9 & 26.7 & 43.8 & 63.3 & 95.4 & 52.5 \\
    \midrule
    \multicolumn{7}{@{}l}{\textit{gpt-oss-20b (video tool served by Gemini 3.1 Flash Lite)}} \\
    Reports only & 32.8 & 26.7 & 43.8 & 33.9 & 30.0 & 37.5 \\
    EHR only     & 37.3 & 26.1 & 27.5 & 23.4 & 93.1 & 27.5 \\
    Video only   & 28.4 & 28.5 & 43.8 & 32.3 & 14.6 & 12.5 \\
    All tools    & 57.4 & 32.1 & 41.2 & 62.5 & 93.1 & 40.0 \\
    \bottomrule
  \end{tabular}
\end{table}

\clearpage
\subsection{Delirium and Atomic Action Frequency Analysis}
\label{app:delirium-analysis}

We define the \emph{delirium frequency ratio} as being:
\begin{equation}
    \mathrm{FR}_a = \frac{f_a^{\mathrm{del}}}{f_a^{\mathrm{non\text{-}del}}},
    \qquad
    f_a^{(g)} = \frac{\sum_{p \in \mathcal{P}_g} m_a^{(p)}}{\sum_{p \in \mathcal{P}_g} M^{(p)}},
    \label{eq:freq-ratio}
\end{equation}
where $\mathcal{P}_g$ denotes the set of patients in group
$g \in \{\mathrm{del}, \mathrm{non\text{-}del}\}$, $m_a^{(p)}$ is the number of
patient-minutes in which action $a$ is detected for patient $p$, and $M^{(p)}$
is the total number of observed minutes for that patient; group frequencies are
thus pooled over all observed minutes within each group (minute-weighted).

The delirium frequency ratios reported in Figure~\ref{fig:atomic-action-deepdive}(b) and Eq.~\ref{eq:freq-ratio} are cohort-level associations, and several features of their construction warrant elaboration.

\textit{Group composition.} Patients who screen positive for delirium may differ from those who do not along dimensions that independently suppress observable movement, including sedation depth, exposure to mechanical ventilation, and length of stay. Table~\ref{tab:delirium-cohort} summarizes these characteristics by group. Ventilation exposure is similar between groups (invasive ventilation 58.5\% vs.\ 55.5\%), as is documented sedation level (per-patient median RASS 0.00 in both groups). The delirium group has a higher proportion of assessments indicating deep sedation (RASS $\leq -3$: 5.0\% vs.\ 2.0\%) and a longer ICU stay (median 4.62 vs.\ 3.09 days), and is correspondingly observed for more minutes. 

\textit{Weighting.} Equation~\ref{eq:freq-ratio} pools numerator and denominator counts across all patients in a group, so each patient contributes in proportion to the number of minutes observed. Patients with longer stays therefore exert greater influence on group frequencies, and length of stay is itself associated with delirium. As a sensitivity analysis, we recomputed a different ratio that weights every patient equally---that is, by first computing a per-patient frequency $m_a^{(p)}/M^{(p)}$ and then averaging across patients within each group (Table~\ref{tab:delirium-fr}). The three largest reductions are essentially unchanged under equal weighting: using non-medical objects ($0.41\times \rightarrow 0.39\times$), leg ADL movement ($0.46\times \rightarrow 0.47\times$), and object exchange ($0.48\times \rightarrow 0.47\times$). For this analysis, several ratios closer to unity became more frequent for delirious patients---adjusting bedding ($0.85\times \rightarrow 1.07\times$), position change ($0.85\times \rightarrow 1.06\times$), and in-bed leg movement ($1.01\times \rightarrow 1.30\times$). However, all of these show minimal differences between groups.

\textit{Uncertainty.} To quantify sampling variability, we computed 95\% confidence intervals for each $\log_2 \mathrm{FR}_a$ by resampling patients with replacement within each group (10,000 bootstrap replicates, patient-level resampling).  Intervals are reported in Table~\ref{tab:delirium-fr}. Of the 13 evaluated actions, seven have intervals excluding zero under \emph{both} weightings and are reduced in the delirium group: using non-medical objects, leg ADL movement, object exchange, gesticulating, other arm movements, grabbing medical equipment, and head turning. One action, clinician--patient contact, is elevated in the delirium group under both the minute-weighted and patient-weighted estimators ($1.20\times$ and $1.18\times$). The remaining five---adjusting bedding, position change, looking at others, in-bed leg movement, and communication---are not distinguishable from no difference under either weighting, and the corresponding bars in Figure~\ref{fig:atomic-action-deepdive}(b) should not be interpreted as evidence of a group difference. Ratios for infrequent actions and for categories with lower detection F1 carry correspondingly wider intervals and warrant greater caution.

\textit{Significance tests.} The bootstrap intervals above cannot resolve small $p$-values, so Table~\ref{tab:delirium-tests} reports significance tests on the same cohort. For each action we compare per-patient rates (minutes with the action over observed minutes) between the 53 patients with a positive screen and the 227 without. We use a two-sided Mann--Whitney $U$ test (tie-corrected asymptotic $p$, since rates are heavily tied at zero), adjusted across the 13 evaluated actions by Holm and by Benjamini--Hochberg. We test each $\log_2\mathrm{FR}_a$ by permuting the group labels 2{,}000{,}000 times (two-sided, $p=(n_{\geq}+1)/(B+1)$, floor $5\times10^{-7}$). We define \emph{purposeful activity} as the per-patient fraction of observed minutes containing any of the following four purposeful actions: using non-medical objects, object exchange, leg ADL movement, gesticulating. We find a median of 0.053 (IQR 0.024--0.118) in the delirium group against 0.146 (0.093--0.217) in the non-delirium group, a pooled rate of 0.080 against 0.161 ($\log_2\mathrm{FR}=-1.01$), with $U=2{,}998$, $p=1.3\times10^{-8}$ by Mann--Whitney and $p=5\times10^{-7}$ by permutation on $\log_2\mathrm{FR}$. The three purposeful actions with the largest reductions remain significant at $p<10^{-4}$ after Holm correction; gesticulating does not ($p=1.5\times10^{-3}$ adjusted). Two actions in the bootstrap-significant set are weaker under these tests: head turning is not significant on per-patient rates ($p=0.094$ unadjusted), and grabbing medical equipment survives Holm correction only narrowly ($p=0.034$). Clinician--patient contact, the one action more frequent in the delirium group, is robust ($p=1.4\times10^{-5}$, Holm $1.4\times10^{-4}$).

\textit{Label timing.} Patients are assigned to the delirium group if any CAM-ICU assessment during the stay is positive, whereas behavior frequencies are pooled over all observed minutes of the stay. The analysis therefore contrasts patients who experience delirium at some point against those with no documented positive screen, rather than contrasting delirious against non-delirious periods within a stay. Analyses aligned to assessment times---for example, restricting to windows adjacent to a positive screen, or comparing CAM-positive against CAM-negative days within the same patient---would more directly probe the motoric phenomenology of delirium and are a direction for future work.

\begin{table}[h]
  \centering
  \footnotesize
  \caption{\textbf{Significance tests for delirium frequency differences.} Per-patient rates (median) by group with two-sided Mann--Whitney $U$ $p$ and Holm and Benjamini--Hochberg adjustment over the 13 evaluated actions. Permutation $p$ for the minute-weighted and equal-weighted $\log_2$ frequency ratio over 2{,}000{,}000 label permutations (floor $5\times10^{-7}$).}
  \label{tab:delirium-tests}
  \vspace{1mm}
  \resizebox{\linewidth}{!}{%
  \begin{tabular}{lcccccc}
    \toprule
    Action & Med.\ del. & Med.\ non-del. & MWU $p$ & Holm & BH & Perm.\ $p$ (mw / ew) \\
    \midrule
    Using non-medical objects  & 0.016 & 0.080 & $1.8\times10^{-10}$ & $2.3\times10^{-9}$ & $2.3\times10^{-9}$ & $5\times10^{-7}$ / $5\times10^{-7}$ \\
    Object exchange            & 0.004 & 0.010 & $8.9\times10^{-10}$ & $1.1\times10^{-8}$ & $5.8\times10^{-9}$ & $5\times10^{-7}$ / $5\times10^{-7}$ \\
    Leg ADL movement           & 0.002 & 0.007 & $7.3\times10^{-7}$  & $8.0\times10^{-6}$ & $3.2\times10^{-6}$ & $4\times10^{-6}$ / $5\times10^{-7}$ \\
    Clinician--patient contact & 0.423 & 0.349 & $1.4\times10^{-5}$  & $1.4\times10^{-4}$ & $4.7\times10^{-5}$ & $1.1\times10^{-3}$ / $1.0\times10^{-4}$ \\
    Other arm movements        & 0.165 & 0.243 & $2.9\times10^{-5}$  & $2.6\times10^{-4}$ & $7.4\times10^{-5}$ & $6.5\times10^{-6}$ / $2\times10^{-6}$ \\
    Gesticulating              & 0.030 & 0.053 & $1.9\times10^{-4}$  & $1.5\times10^{-3}$ & $4.1\times10^{-4}$ & $1.4\times10^{-4}$ / $3.5\times10^{-4}$ \\
    Grabbing medical equipment & 0.081 & 0.102 & $4.8\times10^{-3}$  & 0.034 & $8.9\times10^{-3}$ & $1.1\times10^{-3}$ / $4.1\times10^{-3}$ \\
    Head turning               & 0.363 & 0.399 & 0.094 & 0.56 & 0.15 & $4.3\times10^{-3}$ / $8.6\times10^{-3}$ \\
    Position change            & 0.027 & 0.029 & 0.11  & 0.56 & 0.16 & 0.21 / 0.59 \\
    In-bed leg movement        & 0.022 & 0.019 & 0.23  & 0.93 & 0.29 & 0.95 / 0.028 \\
    Adjusting bedding          & 0.037 & 0.038 & 0.25  & 0.93 & 0.29 & 0.23 / 0.52 \\
    Communication              & 0.240 & 0.257 & 0.27  & 0.93 & 0.29 & 0.080 / 0.20 \\
    Looking at others          & 0.330 & 0.320 & 0.74  & 0.93 & 0.74 & 0.54 / 1.0 \\
    \midrule
    Purposeful activity (any of four) & 0.053 & 0.146 & $1.3\times10^{-8}$ & -- & -- & $5\times10^{-7}$ \\
    \bottomrule
  \end{tabular}}
\end{table}

\begin{table}[h]
  \centering
  \footnotesize
  \caption{\textbf{Cohort characteristics by delirium status.} Delirium is defined as any positive CAM-ICU assessment during the ICU stay.  RASS percentages are computed over all recorded assessments in each group rather than over patients. Observed minutes run from arrival in the monitored room and can exceed ICU LOS, which counts only time at the critical-care level of care.}

  \label{tab:delirium-cohort}
  \vspace{1mm}
  \begin{tabular}{lcc}
    \toprule
    Characteristic & Delirium ($n = 53$) & Non-delirium ($n = 227$) \\
    \midrule
    ICU LOS, days, median [IQR]              & 4.62 [3.01, 8.40]      & 3.09 [1.89, 4.84] \\
    Observed minutes, median [IQR]           & 7{,}560 [4{,}680, 12{,}840] & 4{,}920 [2{,}580, 7{,}440] \\
    Invasive mechanical ventilation, $n$ (\%) & 31 (58.5)              & 126 (55.5) \\
    Per-patient median RASS, median [IQR]    & 0.00 [$-1.00$, 0.00]   & 0.00 [$-0.50$, 0.00] \\
    RASS assessments $\leq -3$, \%           & 5.0 (of 6{,}979)       & 2.0 (of 18{,}402) \\
    \bottomrule
  \end{tabular}
\end{table}

\begin{table}[h]
  \centering
  \footnotesize
  \caption{\textbf{Delirium frequency ratios under weighting schemes with bootstrap intervals.} Minute-weighted ratios follow Eq.~\ref{eq:freq-ratio} and reproduce Figure~\ref{fig:atomic-action-deepdive}(b); equal-weighted ratios average per-patient rates within each group. Intervals are 95\% percentile bootstrap intervals over 10,000 patient-level resamples. Values below $1.00$ ($\log_2$ below zero) indicate actions detected less frequently among patients with documented delirium. The final two actions are excluded from the 13-action evaluation set and from Figure~\ref{fig:atomic-action-deepdive}(b) for lack of sufficient positive instances in the annotated benchmark.}
  \label{tab:delirium-fr}
  \vspace{1mm}
  \begin{tabular}{lcccc}
    \toprule
    & \multicolumn{2}{c}{Minute-weighted} & \multicolumn{2}{c}{Equal patient weight} \\
    \cmidrule(lr){2-3}\cmidrule(lr){4-5}
    Action & FR & $\log_2$ FR [95\% CI] & FR & $\log_2$ FR [95\% CI] \\
    \midrule
    Using non-medical objects   & 0.41 & $-1.28$ $[-1.92, -0.75]$ & 0.39 & $-1.35$ $[-1.94, -0.87]$ \\
    Leg ADL movement            & 0.46 & $-1.13$ $[-1.69, -0.67]$ & 0.47 & $-1.07$ $[-1.59, -0.64]$ \\
    Object exchange             & 0.48 & $-1.05$ $[-1.57, -0.61]$ & 0.47 & $-1.08$ $[-1.55, -0.69]$ \\
    Gesticulating               & 0.60 & $-0.74$ $[-1.17, -0.35]$ & 0.68 & $-0.56$ $[-0.97, -0.21]$ \\
    Other arm movements         & 0.62 & $-0.70$ $[-1.00, -0.42]$ & 0.69 & $-0.53$ $[-0.79, -0.29]$ \\
    Grabbing medical equipment  & 0.67 & $-0.58$ $[-0.87, -0.30]$ & 0.80 & $-0.33$ $[-0.58, -0.09]$ \\
    Head turning                & 0.85 & $-0.24$ $[-0.43, -0.07]$ & 0.90 & $-0.15$ $[-0.30, -0.01]$ \\
    \addlinespace
    Adjusting bedding           & 0.85 & $-0.23$ $[-0.58, 0.14]$  & 1.07 & $0.10$ $[-0.31, 0.49]$ \\
    Position change             & 0.85 & $-0.24$ $[-0.61, 0.14]$  & 1.06 & $0.09$ $[-0.37, 0.52]$ \\
    Looking at others           & 0.96 & $-0.06$ $[-0.25, 0.12]$  & 1.00 & $-0.00$ $[-0.17, 0.15]$ \\
    In-bed leg movement         & 1.01 & $0.01$ $[-0.38, 0.42]$   & 1.30 & $0.38$ $[-0.06, 0.82]$ \\
    Communication               & 0.88 & $-0.18$ $[-0.42, 0.04]$  & 0.94 & $-0.10$ $[-0.30, 0.09]$ \\
    \addlinespace
    Clinician--patient contact  & 1.20 & $0.26$ $[0.14, 0.39]$    & 1.18 & $0.24$ $[0.14, 0.35]$ \\
    \midrule
    \multicolumn{5}{l}{\textit{Not in evaluation set (detection performance not validated)}} \\
    Other torso movement        & 0.90 & $-0.16$ $[-0.46, 0.15]$  & 1.08 & $0.10$ $[-0.25, 0.45]$ \\
        Clinician restraining       & 1.54 & $0.62$ $[0.15, 1.07]$    & 1.33 & $0.41$ $[0.01, 0.79]$ \\
    \bottomrule
  \end{tabular}
\end{table}

\newpage
\subsection{Ethics, Data, and Code Availability}
\label{app:ethics}
\paragraph{Human subjects review.}
This study was approved by \siteirb. All video recordings and linked electronic health record data were stored and processed on secure servers, with access restricted to IRB-approved study personnel.

\paragraph{Data and code availability.}
The continuous video recordings and linked electronic health records used in this study contain protected health information and cannot be made publicly available due to patient privacy regulations and the terms of our IRB approval. Model prompts, the atomic action taxonomy, inference code, and evaluation code are available at \url{https://github.com/zanedurante/tracer-harness}.

\end{document}